\documentclass[pdflatex,sn-mathphys-ay]{sn-jnl}

\usepackage{graphicx}%
\usepackage{multirow}%
\usepackage{amsmath,amssymb,amsfonts}%
\usepackage{amsthm}%
\usepackage{mathrsfs}%
\usepackage[title]{appendix}%
\usepackage{xcolor}%
\usepackage{textcomp}%
\usepackage{manyfoot}%
\usepackage{booktabs}%
\usepackage{algorithm}%
\usepackage{algorithmicx}%
\usepackage{algpseudocode}%
\usepackage{listings}%
\usepackage{multirow}
\usepackage{lmodern}
\usepackage{fontenc}
\usepackage{stackengine}
\usepackage{todonotes}
\usepackage{textcomp}
\usepackage{stmaryrd}
\newcommand\textsub[1]{\stackengine{-.5ex}{}{\scriptsize#1}{O}{l}{F}{F}{L}}

\begin{document}

\title[Article Title]{CAVERNAUTE : a design and manufacturing pipeline of a rigid but foldable indoor airship aerial system for cave exploration}


\author*[1]{\fnm{Louis} \sur{Catar}}\email{louis.catar.1@ens.etsmtl.ca}

\author[1]{\fnm{Jorge E.} \sur{Salas Gordoniz}}

\author[2]{\fnm{Ilyass} \sur{Tabiai}}

\author[1]{\fnm{David} \sur{St-Onge}}

\affil[1]{\orgdiv{INIT Robots}, \orgname{École de Technologie Supérieure}, \orgaddress{\street{1100 rue Notre-Dame O.}, \city{Montréal}, \postcode{H3C 1K3}, \state{QC}, \country{Canada}}}

\affil[2]{\orgdiv{LIPEC}, \orgname{École de Technologie Supérieure}, \orgaddress{\street{1100 rue Notre-Dame O.}, \city{Montréal}, \postcode{H3C 1K3}, \state{QC}, \country{Canada}}}


\abstract{Airships, best recognized for their unique quality of payload/energy ratio, present a fascinating challenge for the field of engineering. Their construction and operation require a delicate balance of materials and rules, making them a compelling object of study. They embody a distinct intersection of physics, design, and innovation, offering a wide array of possibilities for future transportation and exploration. Thanks to their long-flight endurance, they are suited for long-term missions. To operate in complex environments such as indoor cluttered spaces, their membrane and mechatronics need to be protected from impacts. This paper presents a new indoor airship design inspired by origami and the Kresling pattern. The airship structure combines a carbon fiber exoskeleton and UV resin micro-lattices for shock absorption. Our design strengthens the robot while granting the ability to access narrow spaces by folding the structure - up to a volume expansion ratio of 19.8. To optimize the numerous parameters of the airship, we present a pipeline for design, manufacture, and assembly. It takes into account manufacturing constraints, dimensions of the target deployment area, and aerostatics, allowing for easy and quick testing of new configurations. We also present unique features made possible by combining origami with airship design, which reduces the chances of mission-compromising failures. We demonstrate the potential of the design with a complete simulation including an effective control strategy leveraging lightweight mechatronics to optimize flight autonomy in exploration missions of unstructured environments.}




\maketitle

\section{Introduction}\label{sec:intro}

The demand for underground Unmanned Aircraft Systems (UAS) is increasing. Aging infrastructures require inspection~\citep{zhang_unmanned_2023}, glacier geologic features need exploration to study their growth mechanisms~\citep{das_eastern_2019}, and Search \& Rescue (S\&R) operations aim to protect rescue teams~\citep{bogue_disaster_2019}. Ground robots struggle to manage the complex terrains' topology, crawling and rolling robots are slow solutions that are less adapted to large or unstructured areas, making multicopters aerial vehicles a popular option. However, these vehicles face significant autonomy challenges. Commercial solutions designed for exploration and data collection in complex indoor or subterranean environments, such as the Elios 3 from Flyability~\citep{adrien_elios_2024}, cannot fly for more than 13 minutes. Their autonomy decreases further with inspection payloads like LiDAR. Similarly, vehicles developed in university laboratories have flight times that do not exceed 5 minutes~\citep{briod_collision-resilient_2014, falanga_foldable_2019} or 15 minutes~\citep{kratky_autonomous_2021}. These limited flight durations restrict mission outcomes. In this paper, we focus on the use of airship UAS. With these vehicles, power is almost exclusively used for movement, while the lighter-than-air gas provides lift.

Indoor airships have already been proposed for underground applications~\citep{wang_heterogeneous_2022}. The flight longevity offered by buoyancy is a game changer for long missions in complex environments. Deployment in confined spaces (such as the DARPA Subterranean Challenge) is not an obstacle for this type of aerial system, though its size must adapt to narrow passages~\citep{huang_duckiefloat_2019, lu_heterogeneous_2022, hudson_heterogeneous_2022}. With these challenges in mind, we propose a new indoor airship architecture.

Most examples of indoor airships are soft blimps supporting a gondola of actuators and mechatronics. A blimp is an airship whose shape is maintained by the pressure of the gases within its envelope~\citep{hecks_pressure_1972}. For small airships, the ratio of the lift provided by the gas (volume) to the mass of the envelope (surface) is generally not sufficient to support heavier structures. Nevertheless, rigid airships can provide increased reliability and durability, which are critical for harsh environment missions. A rigid configuration can help to reduce drag, increase flight speed, improve balance and control by maintaining an aerodynamic shape, and provide greater design flexibility for the position of the propulsion system.~\citep{konstantinov_basics_2003}.
While the surface/volume ratio favors the sphere, its excessive drag is incompatible with operations in outdoor spaces. Therefore, streamlined shapes offer good aerodynamic performance that only a rigid or semi-rigid structure can ensure. Indeed, soft blimps require increased gas pressure on the envelope to maintain their shape, which increases the weight. Conversely, in indoor spaces, fewer external aerodynamic disturbances affect the flight and shape of the airship, allowing for innovative rigid structures to be explored and created. 

Knowing these ratio constraints, another problematic point is accessibility to remote and narrow interior areas. These large structures need to be transported and pa for field missions. This work introduces a rigid but foldable autonomous airship UAS named CAVERNAUTE (Compact Airship Vehicle for Expeditions into Remote Natural Abysses and Underground Tunnels) for deployment inside a hard-to-reach natural cave. To further push both the geometrical design of an unconventional airship and provide a complete field deployment experience, this project is part of an artistic performance. The CAVERNAUTE is expected to explore a restricted-access cave while broadcasting its journey into an immersive room open to the public. The constraints of this artistic context are thus integrated into the design presented throughout this paper.

This work contributes to the field of robotic design and exploration in complex environments in three core aspects: \textit{(1)} the development of a rigid foldable geometry by applying origami techniques to airship structures; \textit{(2)} the pipeline of the design of a LTA robot with weight, volume, geometric, and manufacturing intertwined constraints for underground mission; and \textit{(3)} a control adapted in regard of flight-autonomy with ultralight-weight mechatronic limitations.

\section{Related work}

\subsection{Rigid airship designs}


In the vast majority of literature, small indoor airships (less than 3 m in length) are based on a combination of ellipse, circle, and/or parabola geometries like GNVR-type shape \citep{gawale_design_2008, joshi_conceptual_2009,ram_multidisciplinary_2010, adeel_design_2017, mistri_design_2017, biju_design_2017}. Other less conventional shapes still rely on inflated volumes, as they do not have a stiff envelope: the gondola is suspended under the airship, or the structure is semi-rigid with rings placed at sections corresponding to the gondola~\citep{gorjup_low-cost_2020, zheng_user_2021}. Standing out from conventional designs are studies with more complex shapes, such as wing-like profiles, torus, or bio-inspired shapes like the ray (hybrid airship)~\citep{edge_lighter-than-air_2010}. Our work is inspired by a previous experiment involving a cubic-shaped airship with a rigid exoskeleton, named Tryphon, deployed in the La Verna cave in France~\citep{st-onge_voiles_2011}.

From the perspective of rigid structures, small airships have not, to our knowledge, gone beyond semi-rigid designs. The semi-rigid structure does not hold the shape without gas~\citep{liao_review_2009}. Typically, rigidity is achieved by rings surrounding the flexible envelope, allowing for better control of the shape either by restricting expansion due to gas pressure~\citep{sharf_development_2013, mistri_design_2017}, or by tensioning the envelope~\citep{edge_lighter-than-air_2010, st-onge_voiles_2011}. In the latter case, the forces on the structure, made up of carbon fiber beams and sometimes honeycomb sandwich panels, can be very demanding without directly taking up the loads via tension elements (cables), which significantly lighten the structure. In the work of~\cite{anil_design_2016}, the authors detail a wide range of ultralight materials for making a semi-rigid airship, including balsa wood and composite materials.

Exoskeletons can also be adapted to improve the propulsion system. Notably, the work of ~\cite{jordi_body_2008} was inspired by the swimming movement of fish to construct an actuator for an airship using electro-active polymer (EAP) dielectric actuators. The final aircraft was 8 meters long and included three actuators and a structure linking the joints, achieving a top speed of up to 0.45 m/s. They achieved good efficiency with this biomimetic system, but the flight performance led them to conclude that such a propulsion system would be relevant for airships less than 3 meters in length. The applications of this type of silent propulsion are numerous, including wildlife observation and discreet surveillance.

\subsection{Airship design pipeline}

Adopting an approach of standardization and modularization in the design of airships, similar to practices in the rest of the aerospace industry, would offer several key advantages. First, standardization can significantly reduce production and maintenance expenses by enabling the large-scale manufacturing of interchangeable parts, creating economies of scale, and simplifying supply chains~\citep{titenkov_prospects_2018}. As for modularization, it simplifies the complexity of systems by separating functions into distinct modules, which facilitates technological updates and repairs~\citep{larsson_standardization_2018}, but also the design of modules with different set of requirements~\citep{dmitriev_improving_2019}. Nevertheless, the airship's industry is still too young to leverage these advantages, which in turn does not yet provide much incentive for researchers to consider standardization and modularization in their design.

\begin{figure}[h]
\begin {center}
\includegraphics[width=0.45\textwidth]{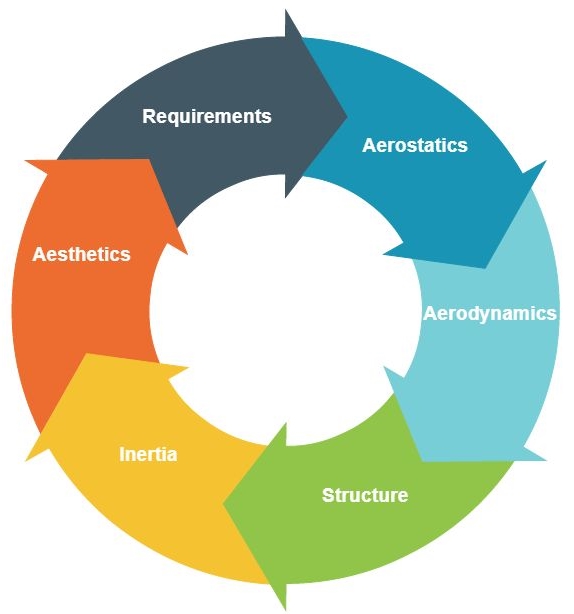}
\caption{Interaction between the different requirements in the case of an aerostat. There is a mutual dependency between the multidisciplinary criteria requiring the use of a design methodology.}
\label{fig:criteria}
\end {center}
\end{figure}

Beyond the shape, only few studies have endeavored to define a framework for designing airships. Numerous airship projects have emerged lately, most with unconventional configurations \citep{ceruti_conceptual_2014}. Due to the disparity of designs, each approach to design does not follow a defined or generalizable framework. However, the airship, with its aerostat characteristics, is composed of the same technical requirements regarding its flight behavior (lift, structure, etc.). They are very intertwined, as shown in Fig.~\ref{fig:criteria}, which does not facilitate the design.

Thus, trying to frame and verify compliance with requirements according to different nested requirements is something that can be introduced throughout the airship design cycle to follow the correct correlation and impacts on the performance of the final model. This methodology, such as the use of genetic algorithms to explore numerous configurations, allows good use of the toolkit offered by understanding airship modeling and relying less on the experience of aircraft designers \citep{liao_review_2009}. \cite{zheng_user_2021} proposed already a modular design framework for the realization of custom airships. However, it remains limited to geometry with one single parameter (diameter) and conventional GNVR-type envelope shape~\citep{suvarna_design_2019}. 

Airships lend themselves well to long-lasting missions. Thanks to their autonomy and a lower cruising speed than a multicopters, their deployment is perfectly fitted to gradual acquisition of data such as LiDAR scanning. Palossi et al. highlight the relevance of small indoor airships compared to flying wings and multicopters~\citep{palossi_self-sustainability_2017}. They specify their uniqueness in terms of maneuverability, their ability to govern, and agility, i.e., their maneuverability in a restricted space at low speed.
Still, for a practical use case, the payload often requires rather large volumes, such as the one designed by~\cite{salas_gordoniz_modular_2021, salas_gordoniz_scutigera_2022}: a bio-inspired multi-body architecture to move a high-lift aerostat in caves and tunnels. Many control challenges are associated with this type of architecture, as it is radically different from the common ellipsoid, but also large with respect to the cluttered environment targeted for their missions.

Nevertheless, the autonomy of new airship designs is rarely directly addressed in the literature, where more emphasis is on the control challenges. Among the few examples, \cite{cho_autopilot_2017} mention that their blimp, the GT-MAB (volume of 0.12 m\textsuperscript{3}) can fly for up to two hours, while the “Tryphon” cubic airships (volume of 9.85 m\textsuperscript{3}) can fly up to six hours~\citep{st-onge_modelisation_2011}, and~\cite{salas_gordoniz_intuitive_2024} achieved up to 41 minutes flights with their robot (volume of 2.4 m\textsuperscript{3}).


\subsection{Folding airships}

There are many advantages to creating a folding airship. On large aircraft, it helps to reduce drag~\citep{chillara_review_2019, zhang_aerodynamic_2021}, and to design more efficient aerodynamic control surfaces~\citep{cozmei_aerogami_2020}. It also allows being more compact to meet specific mission requirements, such as an airship project for space exploration of Titan, which has to fit into the nose cone of a rocket~\citep{duffner_conceptual_2007}, or the foldable exoskeleton of the multi-body blimp of Salas Gordoniz et al. to better access underground areas~\citep{salas_gordoniz_scutigera_2022}. 

When it comes to creating an airship with rigid but foldable characteristics, the association evokes origami. Origami, a Japanese art that uses folding to create three-dimensional structures, has found extensive applications in engineering due to its unique properties: deployability, scalability, self-actuation, reconfigurability, tunability (capability to be optimized), and ease of manufacturing~\citep{callens_flat_2017, meloni_engineering_2021}. The aerospace field has been particularly prolific in adopting origami techniques, largely due to the constraint of transporting structures in cramped rocket fairings before deployment in space~\citep{zirbel_hanaflex_2015, chen_autonomous_2019, yang_volume_2023}. Additionally, origami has numerous applications in the biomedical field, such as in surgical devices~\citep{suzuki_origami-inspired_2020, banerjee_origami-layer-jamming_2020} and implants~\citep{kuribayashi_self-deployable_2006, kim_hydrogel-laden_2015, bobbert_russian_2020}.

\begin{figure}[H]
\centering
\includegraphics[width=0.7\textwidth]{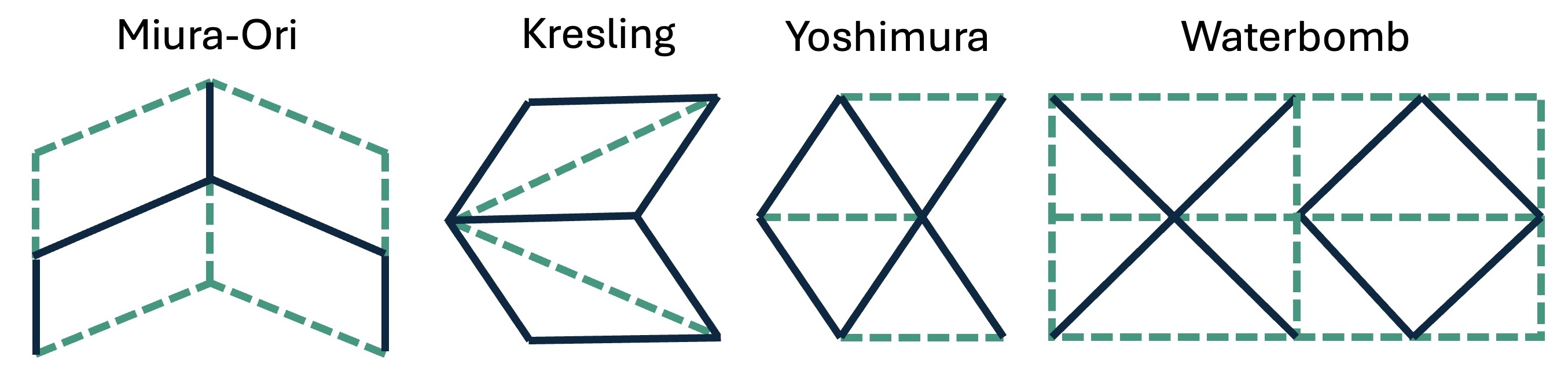}
\caption{Origami folding pattern. Valley creases are shown in dotted green and mountain creases in solid blue. }
\label{fig:pattern}
\end{figure}

Robotics has also benefited from various mechanisms inspired by origami, including those that enable crawling robots~\citep{onal_origami-inspired_2013, pagano_crawling_2017, lee_highload_2021} and create complex mechanisms~\citep{kim_origami-inspired_2018, novelino_untethered_2020, zhang_pneumaticcable-driven_2020, zhang_3d_2021}. Among these mechanisms, origami structures with enclosed volumes are particularly suitable for shaping airships. These structures can be categorized into four main families: Miura-Ori, Kresling, Yoshimura, and Waterbomb patterns~\citep{fossati_origami_2022}. The folding patterns of the individual cells of each type are shown in Fig.\ref{fig:pattern}. Among these, the Kresling pattern is peculiar: a biomimetic design that draws inspiration from natural forms like the abdominal bellows of moths\citep{kresling_natural_2008} and phenomena of dynamic shear rupture~\citep{kresling_fifth_2020}.

Origami structures have typically been studied for their kinematics and the forces they can produce. There are some works that consider the thickness of the material for manufacturing~\citep{hwang_effects_2021, li_design_2023}, but studies on the contained volume are sparse~\citep{lynch_volumetric_2020}. Notably, Yang et al. optimized the volume for a space habitat to fit inside a SpaceX Falcon 9 rocket payload fairing using a genetic algorithm, showing that the Miura-Ori pattern had a favorable radial expansion compared to the Kresling pattern~\citep{yang_volume_2023}. However, their study was volume-driven, and more research is required to include additional manufacturing constraints and mission requirements.


\section{CAVERNAUTE design}

Full surface coverage with rigid structure, even ultra-lightweight such as micro-lattices, are too heavy to be considered as a realist solution.
In this section, we will first introduce an origami pattern fit for airship design. We will then detail the elements that compose the structure of the CAVERNAUTE before presenting the parametric design pipeline that encompasses the geometry, manufacturing, and assembly requirements of the aerial vehicle.

\subsection{Kresling-inspired structure}\label{sec:kresling}

\begin{figure}[H]
\centering
\includegraphics[width=0.7\textwidth]{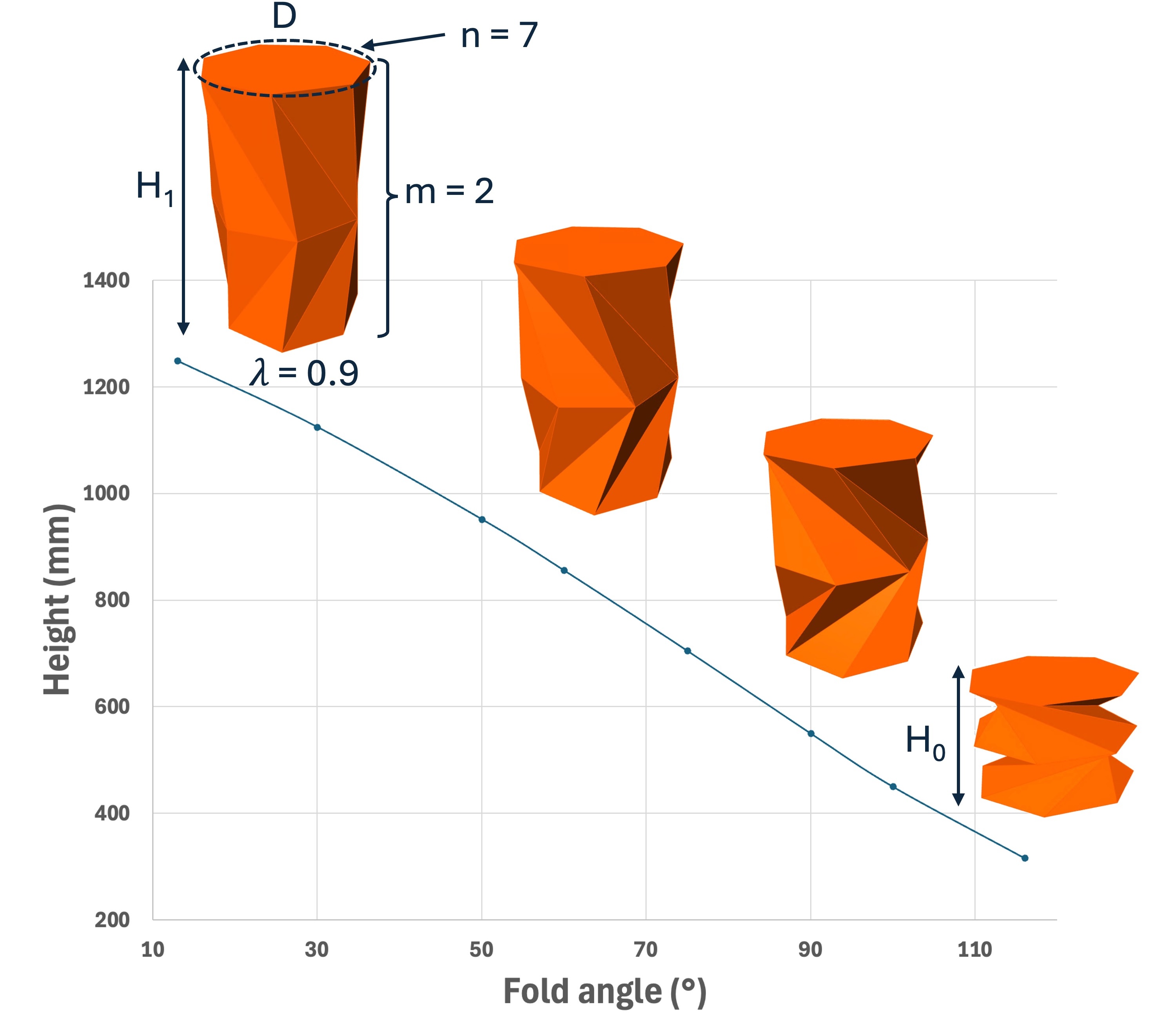}
\caption{Folding motion of a Kresling pattern ($n = 7,~m=2,~\lambda=0.9$). Folding follows a shearing movement. Due to segment symmetry, there is only pure translation for each bi-segment. }
\label{fig:folding}
\end{figure}

Kresling pattern is selected because of the low number of edges required to create a cylinder compared with other patterns. It is composed of congruent parallelograms containing diagonal folds that form a cylinder when the mountains and valleys are folded (Fig.~\ref{fig:folding}). Each stage constitutes a segment. A Kresling cylinder can be described using 6 independent parameters: H\textsub{1} (unfolded height), H\textsub{0} (folded height), D (external cylinder diameter), n (sides number of polygon), m (number of segments) and $\lambda$ (one angle ratio). An example is shown in Fig.~\ref{fig:folding}.

The notations and equations are taken from the geometric description of a Kresling segment done by \cite{bhovad_using_2018, bhovad_peristaltic_2019}. 

\begin{figure}[H]
\centering
\includegraphics[width=1\textwidth]{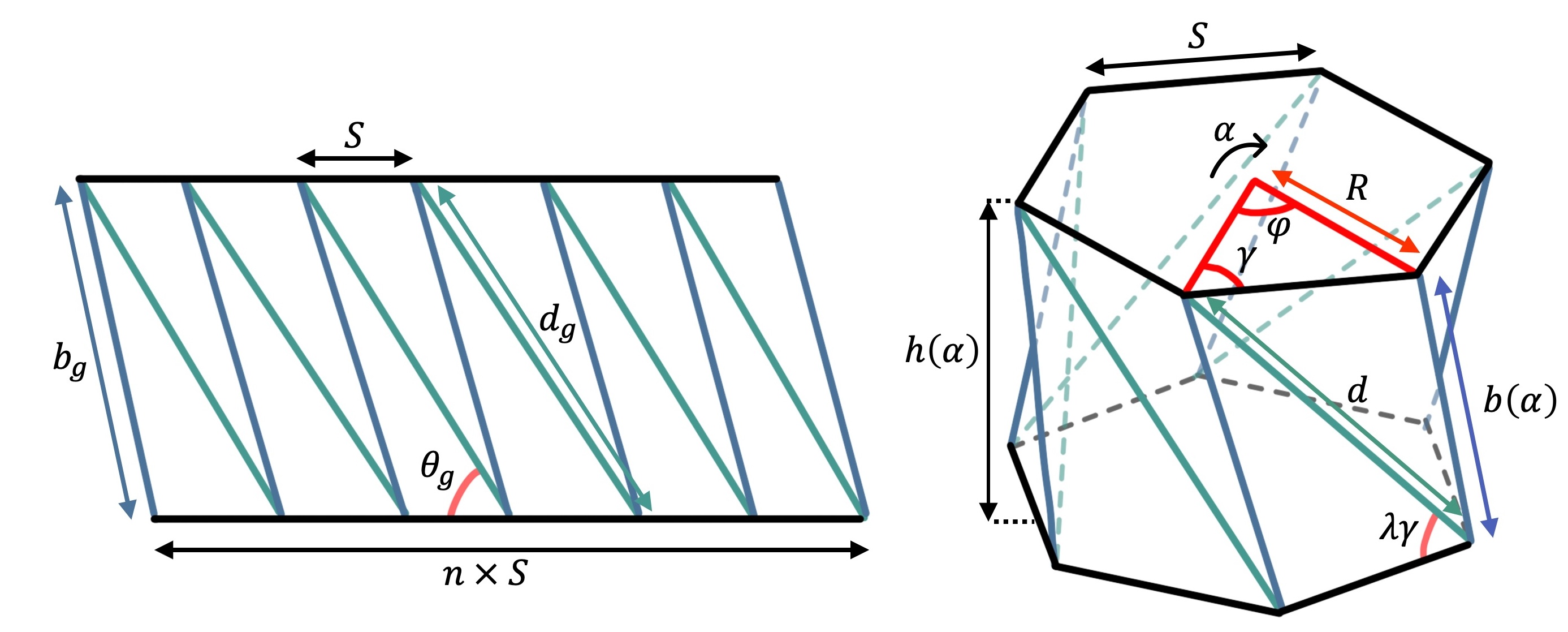}
\caption{Kresling geometrical parameters of one segment ($n = 6,~m=1$). $\alpha$ is the angle of rotation to unfold the geometry between the folded \textit{(0)} and unfolded \textit{(1)} state. The valleys are in green and the mountains in blue. $\alpha_{(0)} = 2 \lambda \gamma$ and $\alpha_{(1)} = 2(1-\lambda)\gamma~\forall \lambda > 0.5$}.
\label{fig:kresling}
\end{figure}

The schematic diagram of a Kresling segment is shown in Fig.~\ref{fig:kresling}. The opening angle $\varphi = \frac{\pi}{n}$ defines the opposite angle $\gamma = \frac{\pi}{2}-\varphi$ and the side length $s=2~R~sin(\varphi)$. The angle ratio $\lambda$, is the ratio of the angle of the diagonal crease to half the internal angle of the base polygon $\gamma$. From the two classical parameters of side $b_c = \sqrt{s^2+{d_c}^2-2~s~d_c~cos(\lambda\gamma)}$ and diagonal $d_c = 2~R~cos(\gamma - \lambda\gamma)$, we can describe generalized parameters that consider the thickness of the material with a minimum bent height $h_0$: $b_g = \sqrt{{b_c}^2+{h_0}^2}$, $d_g = \sqrt{{d_c}^2+{h_0}^2}$ as well as the angle $\theta_g = cos^{-1} \left(\frac{s^2 + {d_g}^2 - {b_g}^2}{2~s~d_g}\right)$. These three generalized parameters define the shape of the Kresling segment envelope. For the opening kinematics of the Kresling pattern, only two parameters (height $h$ and $b$) vary according to the $\alpha$ opening angle. These are described in correspondent equations~\ref{eq:halpha} and~\ref{eq:balpha} and Fig.~\ref{fig:kresling}. 

\begin{equation}
    h(\alpha) = \sqrt{{h_0}^2+2~R^2~[cos(\alpha + 2\varphi)-cos(\alpha_0 + 2\varphi)]}
    \label{eq:halpha}
\end{equation}

\begin{equation}
    b(\alpha) = \sqrt{2~R^2~(1-cos(\alpha))+h(\alpha)^2)}
    \label{eq:balpha}
\end{equation}

\begin{figure}[H]
\begin{minipage}[l]{.48\textwidth}
    {\includegraphics[width=\textwidth]{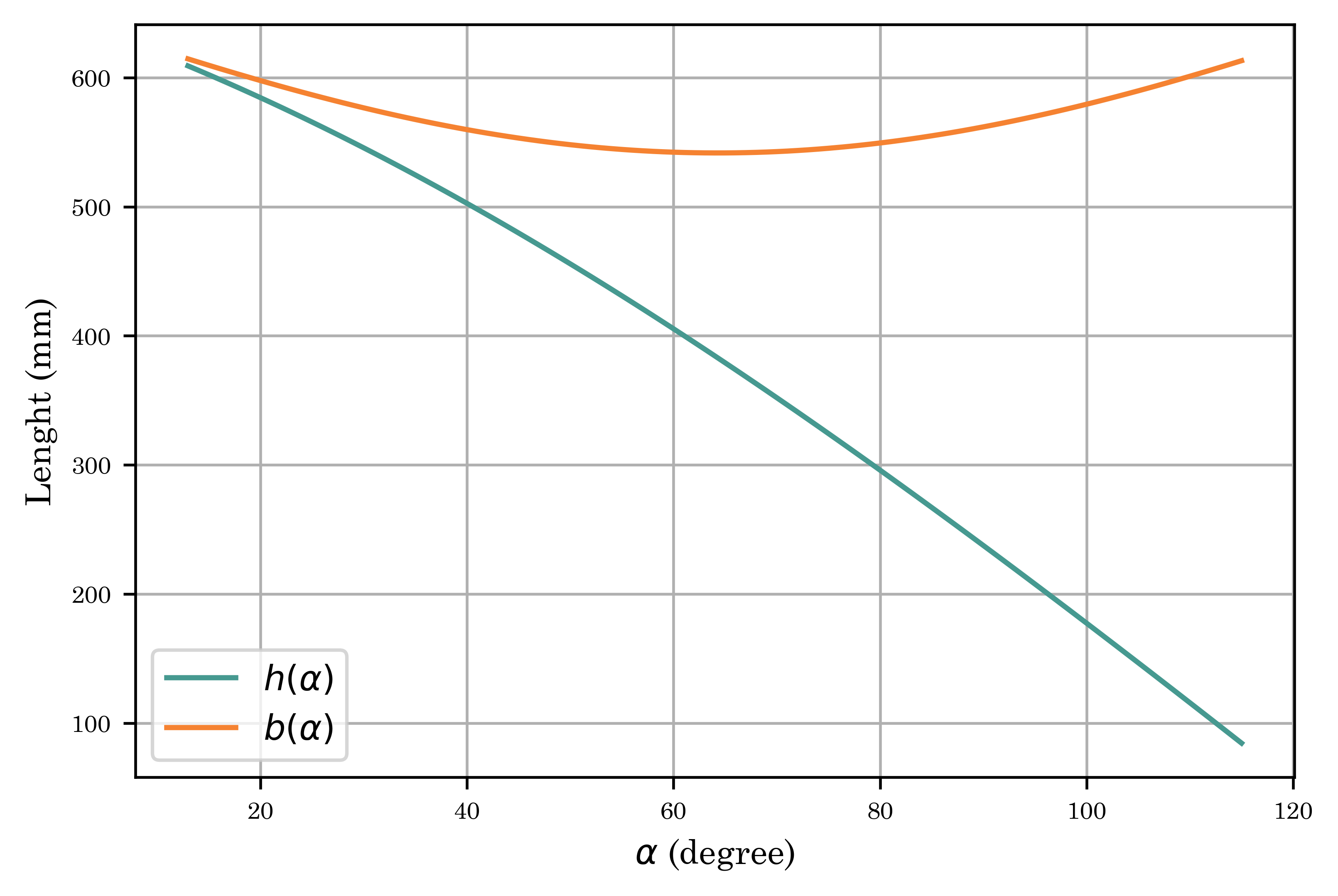}}
    \label{fig:hb}
    \caption{Evolution of the length of $h$ and $b$ as a function of the bending angle $\alpha$. With the geometric description, the diagonal $d_g$ and the sides of polygons $s$ do not vary as a function of $\alpha$. However, this variation in $b$ creates a constraint that can be represented in Fig.~\ref{fig:energy}}
\end{minipage}
\hfill    
\begin{minipage}[l]{.48\textwidth}
    {\includegraphics[width=\textwidth]{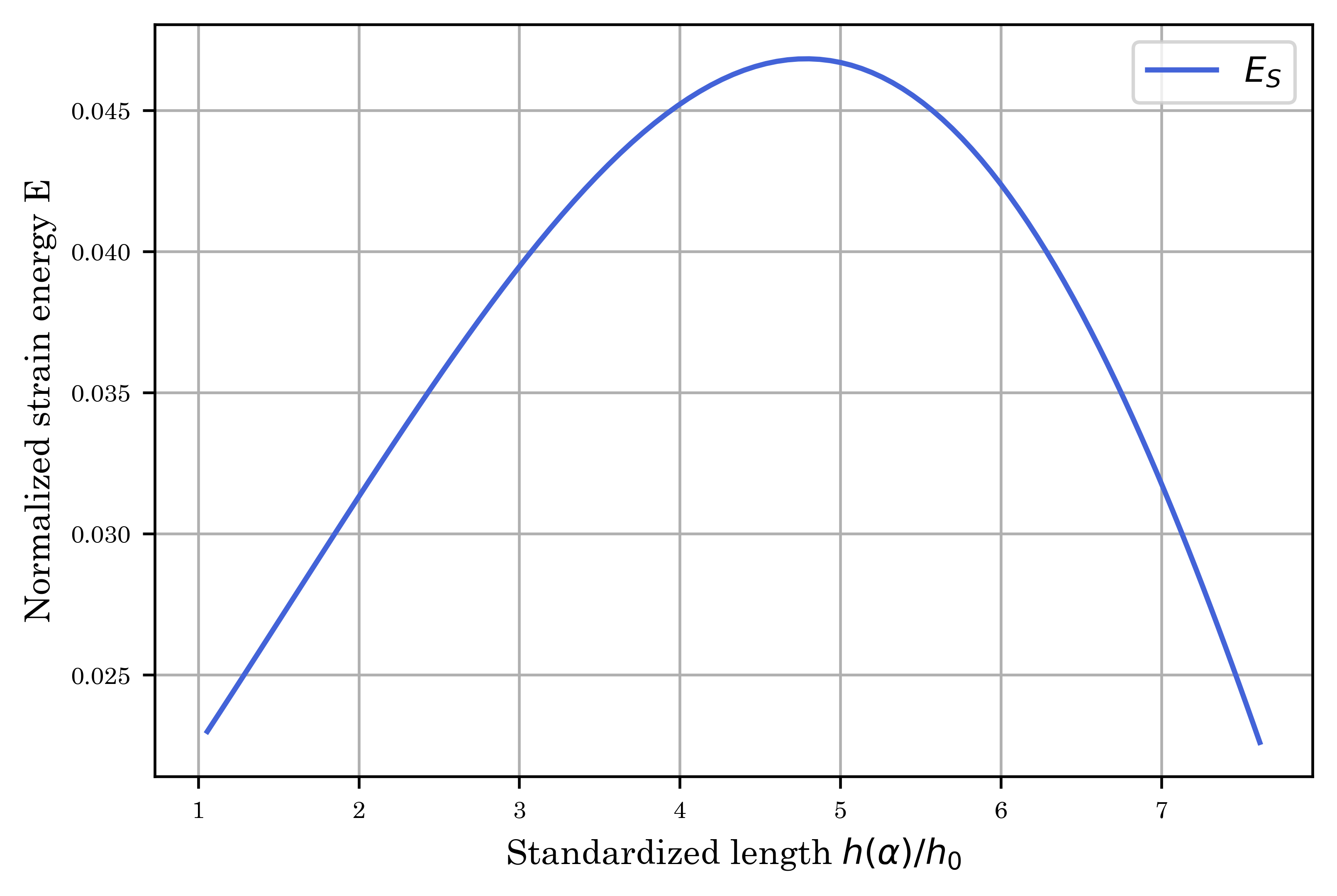}}
    \label{fig:energy}
    \caption{Normalized energy $E_S$ as a function of the relative height of a Kresling segment. This shows the bistability between the folded and unfolded configurations.}
\end{minipage}
\end{figure}

The Kresling pattern has been adjusted so that it is bistable ($\lambda > 0.5$). Figure~\ref{fig:hb} shows that the height $b(\alpha)$ of the segment changes according to the opening of the Kresling.
Equivalent strain $\varepsilon = \frac{b(\alpha)}{B_g} - 1$ and strain energy $U = \frac{1}{2}~K~\varepsilon^2$ can be defined with K, the material stiffness. Then normalized strain energy is computed as: $E_S = \frac{1}{2}~\varepsilon^2$. 
This variation of energy potential induces a bistability of CAVERNAUTE (Fig.~\ref{fig:energy}). In the folded and unfolded position.

\subsection{Exoskeleton design for manufacturing}\label{sec:equation}

The exoskeleton is the rigid structure that maintains the shape of the airship's envelope and ensures the structural coherence of all CAVERNAUTE components, while offering foldable characteristics for compactness. It must be as light as possible. Therefore, the exoskeleton design process focuses on each part separately to leverage the best mechanical performance of each. 

\begin{figure}[H]
\centering
\includegraphics[width=0.8\textwidth]{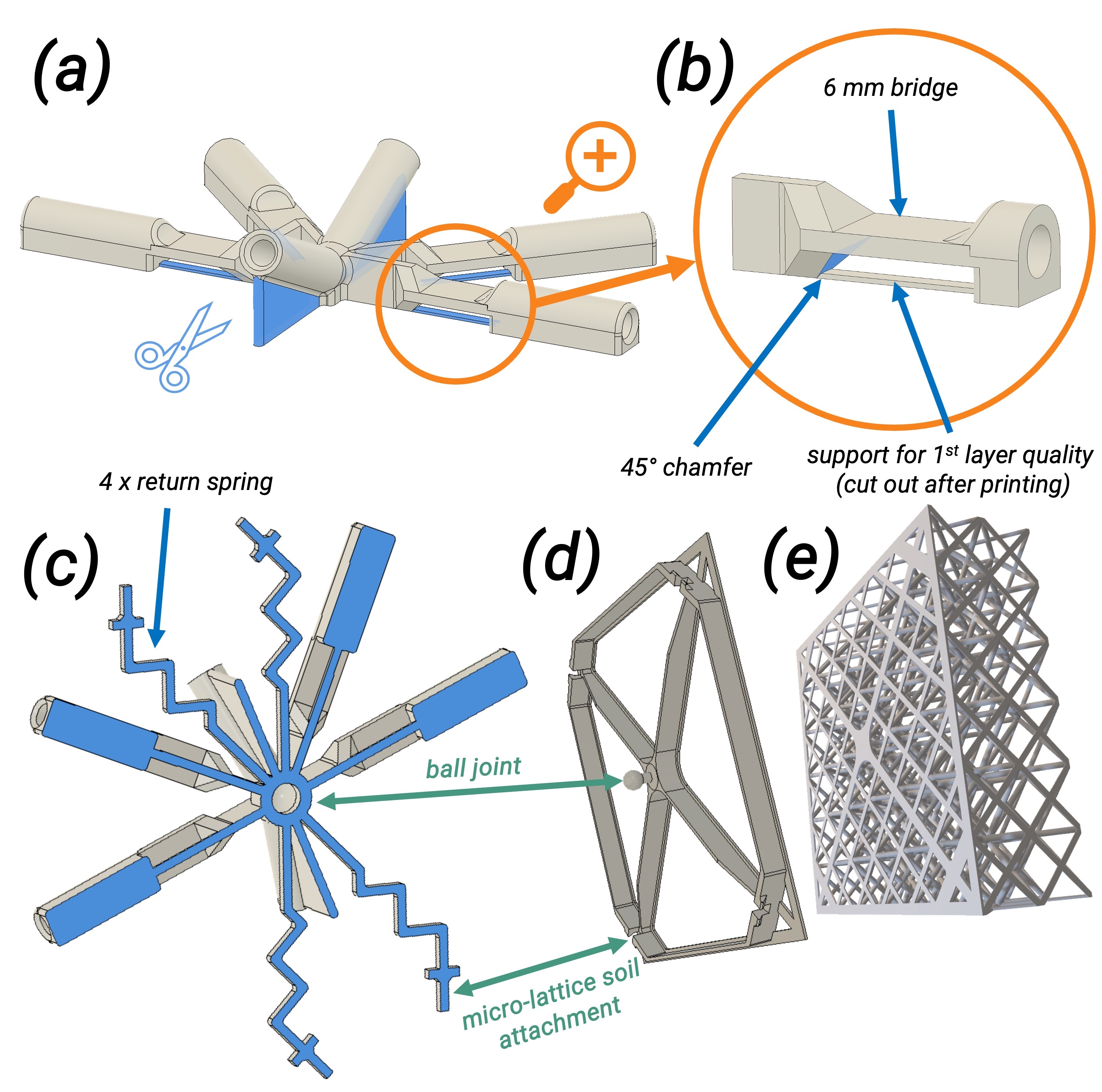}
\caption{Exoskeleton tube junction: (a) simple and (c) to host lattices. Manufacturing features are highlighted in blue. Sub-figure \textit{(c)} shows the micro-lattices junction with the micro-lattices soil \textit{(d)} return spring and ball joint connection. Sub-figure \textit{(e)} shows the micro-lattice buffer structure.}
\label{fig:junction}
\end{figure}

In order to joint the tubes and provide folding capability, 3D-printed parts in TPU95 were created. These parts were specifically designed to minimize mass, utilize the flexibility of the material to create a compliant mechanism, while still easily printable. The TPU junctions are displayed in Fig.~\ref{fig:junction}. All Design for Additive Manufacturing (DfAM) rules were employed, and print parameters were optimized (Fig.~\ref{fig:junction}): to improve the print, the first layer is continuous to facilitate adhesion \textit{(b)}. Sub-figure \textit{(a)} shows the parts removed after printing to help stability during manufacture. Sub-figure \textit{(b)} details the part's overhangs. TPU is a material that tends to ooze during printing, so we benefit from high travel speed features of the printer (500 mm/s) to eliminate this effect. The final iteration of these junctions (simple configuration shown in Fig.~\ref{fig:junction}(a)) have a weight of m\textsub{simpleTPUjct} = 0.75~g each.

At the ends most exposed to the external environment, micro-lattice structures (e) were added to absorb impact energy. The pattern, material, and cell size are derived from a previous characterization work~\citep{catar_polymer_2024}. A wide range of configurations have been evaluated on a test rig specially designed for crash-testing small, light aircraft. In the end, the face-centered-cubic (FCC) pattern is the most suitable configuration in terms of high-specific energy absorption and efficiency for low velocity impact, as well as considered its easiness in terms of manufacturing. 
A special attachment was designed to host these structures. The variant of the TPU junction includes a ball joint and return springs system, as shown in Fig.~\ref{fig:junction} (c) and (d). This mechanism ensures that the micro-lattice is in the optimal frontal position on impact. To interface between the micro-lattice (e), printed in UV resin (MSLA Saturn S), and the TPU junction, a LW-PLA soil was printed (d). This component is clipped onto the ball joint and secured with the ends of the return springs. LW-PLA is a type of PLA polymer with a thermally activated expanding agent, creating porosity that lighten the part by 50\%. The micro-lattice, the LW-PLA soil, and the modified TPU junction have a total weight of m\textsub{latticeeTPUjct} = 3.4~g.

\vspace{3em}

The airship manufacturing requires the complete bill of material for production. The total tube length needed for a given Kresling configuration is computed with:
\begin{equation}
    L_{CVNT} = m \times n \times (d_g + b_g + s) + n \times s,
\end{equation}
where $m$ stands for the number of segments, n the sides number of the polygon. d\textsubscript{g}, b\textsubscript{g} and s are described in Fig.\ref{fig:kresling}.

Then the total envelope surface for that exoskeleton structure is calculated from the following formula (derived from Heron's formula — ~\citep{sangwin_herons_2024}):
\begin{equation}
    S_{CVNT} = \frac{2~s^2~n}{4~tan(\varphi)} + 2~n \sqrt{\frac{s + d + b}{2}\left(\frac{s+d+b}{2}-s \right) \left(\frac{s+d+b}{2}-b \right) \left(\frac{s+d+b}{2}-d \right)}+S_{sheath}
\end{equation}
Since the exoskeleton maintains the envelope of the airship passing through sheath sew to the envelope, these added membrane surface are computed with
\begin{equation}
    S_{sheath} = L_{sheath} \times t_{sheath},
\end{equation}
with $t_{sheath}$ the patches' width and $L_{sheath}$ given by:
\begin{equation}
    L_{sheath} = (L_{CVNT} - d_g \times n \times m) \times \frac{r_{\%}}{100},
\end{equation}
with $r_{\%}$ the ratio between the length of the sheaths and the length of the CAVERNAUTE edges in percentage, $L_{CVNT}$ the total tube length needed for one CAVERNAUTE, and d\textsubscript{g}, n and m defined in Fig.~\ref{fig:kresling}.

\subsection{Parametric design}

To generate viable designs, we created a pipeline to explore various concepts, integrating manufacturing requirements as well as aerostatic and other key considerations into a Grasshopper\textregistered visual script (available online). Grasshopper\textregistered is a visual programming software application designed to be used with the Rhino\textregistered CAD application. We have adapted its traditional use to leverage both the CAD capabilities and its mathematical blocks to evaluate the physics of the airship, automatically test different concepts, and assist in the manufacturing of the CAVERNAUTE.

\begin{figure}[H]
\centering
\includegraphics[width=1\textwidth]{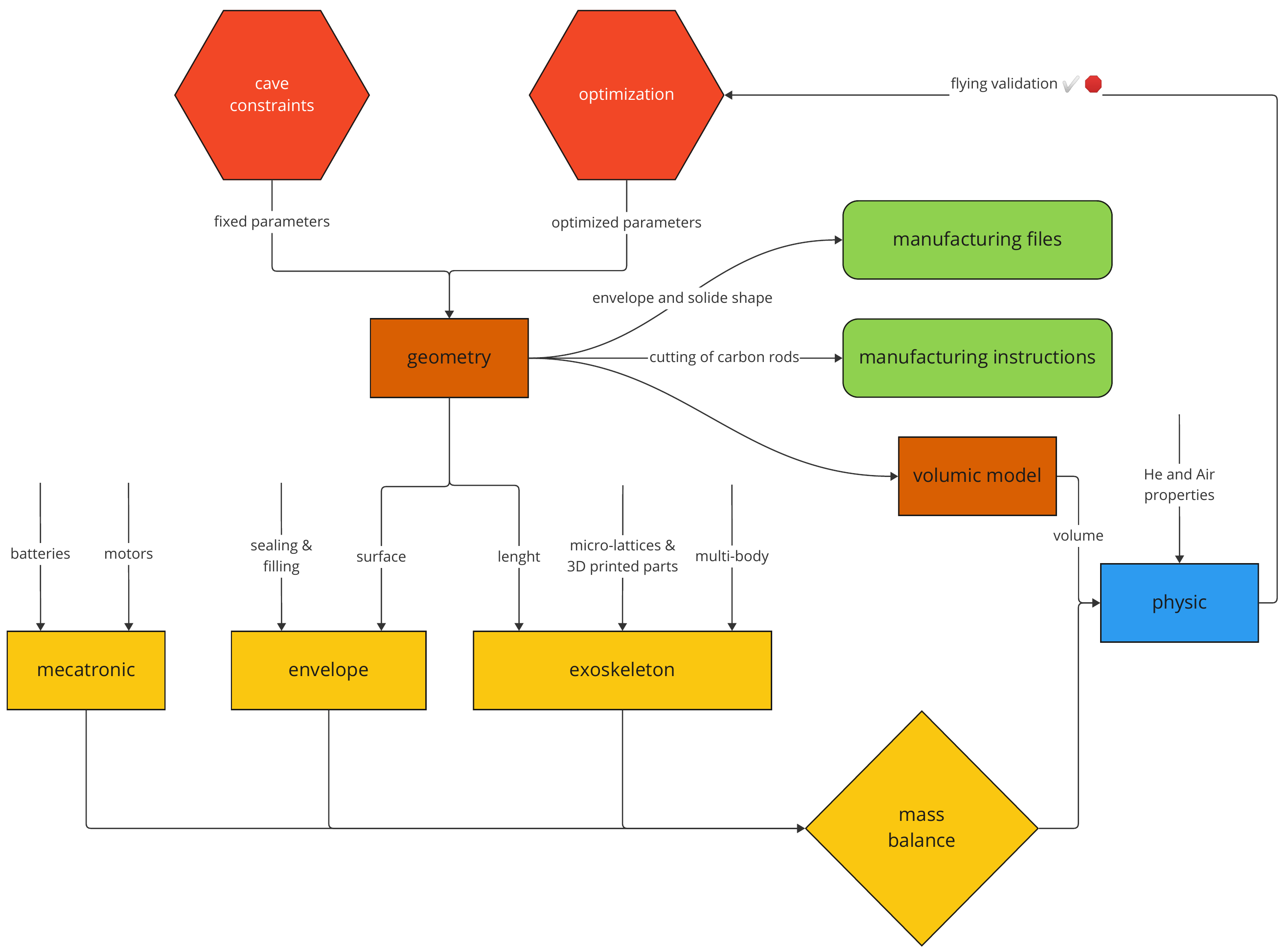}
\caption{Diagram of the CAVERNAUTE design and manufacturing pipeline. The geometric capabilities of the Grasshopper software combined with Rhino are mixed with mathematical capabilities to create a pipeline that considers the many multidisciplinary constraints of a foldable airship design.}
\label{fig:pipeline}
\end{figure}

Figure~\ref{fig:pipeline} shows the flow diagram of our interface. There is a “geometric” group fed by parameter constraints of the target environment (measured) and the three geometric parameters $n$, $m$, and $\lambda$ to optimize. Using the equations presented in Section~\ref{sec:equation}, the pipeline evaluates different characteristic quantities based on the geometry and parameters imported by the user (Table~\ref{tab:number}). The yellow groups allow us to calculate the weight of the onboard components, such as the weight of the exoskeleton or the envelope. From the equations of aerostatics, we can verify buoyancy. A Python script\footnote{\href{https://git.initrobots.ca/lcatar/cavernaute.git}{https://git.initrobots.ca/lcatar/cavernaute.git}} explores all possible variations of the three parameters and records the data for analysis. From the geometry, we represent the volume of the CAVERNAUTE and generate the 2D pattern file for the envelope making. Finally, a last manufacturing module helps with assembly. It provides instructions for the order and lengths of tube cuts.

\vspace{2em}
\vspace{2em}

To feed this pipeline, thirty-two input parameters are editable. Three of them have to be optimized among several thousand possible combinations, others are constant data such as the density of materials, measurements from the target environment and physics as well as some preferences from the user's choices. More advanced customizations are possible, such as the ability to create multiple segments (multi-body airship). Table~\ref{tab:number} groups all these parameters.

\begin{table}[h!]
\centering
\caption{Input parameters editable in Grasshopper, with values used in the current specifications. Some are fixed by the mission, others can be chosen by the user (number of motors, multi-body airship, etc.) and finally, three parameters can be optimized according to the configuration ($n, m, \lambda$).}
\label{tab:number}
\begin{tabular}{cclccc}
\textbf{N°} & \textbf{Parameters} & \multicolumn{1}{c}{\textbf{Description}} & \textbf{Value} & \textbf{Units} & \textbf{Type} \\\midrule
1 & $n$ & number of polygon sides & $\llbracket 3;10 \rrbracket$ & $\mathbb{N}$ & \multirow{3}{*}{\begin{tabular}[c]{@{}c@{}}geometrical \\(to optimize)\end{tabular}} \\
2 & $m$ & number of CAVERNAUTE segment & $\llbracket 2;10 \rrbracket$ & $\mathbb{N}$  \\
3 & $\lambda$ & angle ratio & $[0.50;1.00]$ & n.a. &  \\\midrule
4 & $D$ & CAVERNAUTE diameter & 720 & $mm$ & \multirow{3}{*}{\begin{tabular}[c]{@{}c@{}}geometrical \\(cavern \\ dimensions)\end{tabular}} \\
5 & $H_0$ & CAVERNAUTE folded height & 320  & $mm$ &  \\
6 & $H_1$ & CAVERNAUTE unfolded height  & 2440 & $mm$ &  \\\midrule
7 & $\rho_{air}$ & air density at 12 °C and 99‰ RH & 1.231 & $kg/m^3$ & \multirow{2}{*}{physic} \\
8 & $\rho_{He}$ & helium density & 0.1692 & $kg/m^3$ &  \\\midrule
9 & $m_{CT}$ & carbon tubes linear density & 3.76& $g/m$ & \multirow{21}{*}{manufacturing} \\
10 & $m_{mecatrn}$ & mechatronics weight (board, wires) & 30 & $g$ &  \\
11 & $m_{motors}$ & motor weight & 9.1 & $g$ &  \\
12 & $m_{propellers}$ & propellers weight & 0.46 & $g$ &  \\
13 & $N_{motors}$ & number of motors & 4 & $\mathbb{N}$ &  \\
14 & $m_{battery}$ & battery weight & 80 & $g$ &  \\
15 & $N_{battery}$ & number of batteries & 1 & $\mathbb{N}$ &  \\
16 & $m_{simpleTPUjct}$ & TPU junction weight & 0.75 & $g$ &  \\
17 & $m_{latticeeTPUjct}$ & TPU + micro-lattice patch weight & 3.4 & $g$ &  \\
18 & $N_{patchs}$ & number of micro-lattice patches & 17 & $\mathbb{N}$ &  \\
19 & $d_{env}$ & envelope density & 70 & $g/m^2$ &  \\
20 & $d_{glue}$ & sealing adhesive density or thermic &0& $g/m^2$ &  \\
21 & $t_{ovlp}$ & weld overlap & 10 & $mm$ &  \\
22 & $N_{seal}$ & number of radial sealing lines & 2& $\mathbb{N}$ &  \\
23 & $t_{sheath}$ & sheath width &35& $mm$ &  \\
24 & $r_{\%}$ & sheath ratio &10& \% \\
25 & $m_{valve}$ & valve weight &7& $g$ &  \\
26 & $N_{valve}$ & number of valves &1& $\mathbb{N}$ &  \\
27 & $N_{exo}$ & number of multi-body exoskeletons & 0 & $\mathbb{N}$ &  \\
28 & $N_{CVNT}$ & number of bodies & 1 & $\mathbb{N}$ &  \\
29 & $m_{kevlar}$ & linear density of Kevlar wire &0.28& $g/m$ &  \\
30 & $Dist_{CVNT}$ & distance between multi-bodies &0& $mm$ &  \\
31 & $L_{tubes}$ & raw carbon tube length &1000& $mm$ &  \\
\midrule
 &  &  & 
\end{tabular}
\end{table}

\section{Flight control}

Several conditions are considered when defining the control and flight of CAVERNAUTE. It is intended to fly for several days over multiple weeks, so helium leakage is expected; due to its expansion, it is difficult to maintain a consistent quantity of helium for each flight after re-inflation. These conditions mean that the weight and buoyancy, as well as volume and shape, will not remain exactly constant between flights. Additionally, since CAVERNAUTE is intended to fly in unstructured environments, using a precise model-based controller for a predefined trajectory is impractical.

Attempting to obtain common parameters to identify the model of the robot, such as mass and center of mass, buoyancy and volumetric center, drag estimation, and added/virtual mass/inertia, to then define a flight trajectory, set gains, and conduct several simulated and real-life tests is a tedious task that can jeopardize the deployment of the robot.


To mitigate these problems and ensure safe flight under a generalizable set of conditions, we are using the controller presented in~\cite{salas_gordoniz_intuitive_2024}. This controller is based on two main known characteristics: motor actuation and the velocity optimization required to travel the necessary distances.

The version of the controller we leverage for this work does not require mass/inertia estimation. Instead, the actuation forces and moments ($\overrightarrow{\tau}$) for the inertial axes $X_N, Y_N, Z_N$ and yaw ($\psi$) are defined as follows:
\begin{equation}
    \tau_i
    =F_i\cdot
    sat
    \left(
    \frac{s_i}{\xi_i\cdot tol_i}\right)-
    \xi \dot{i}
\end{equation}
where $i=\{x,y,z,\psi\}$,  
$F_i$ is the maximum force considered for coordinate $i$, $s_i$ corresponds to the sliding surface with slope $-\xi$, same as in Sliding Mode Control (SMC)~\citep{slotine_applied_1991, marquez_nonlinear_2003}, defined as $s_i=\xi_i\tilde{i}+\dot{\tilde{i}}$ In this case, we aim to reach each point of the path at zero velocity, where $\tilde{i} = i - i_d$, and $tol_i$ is the separation from the sliding surface $s_i = 0$, measured in distance units, marking a transition between SMC and PD control. The gain $\xi$ is selected as $F_i / v_{i,max}$ to achieve a cruising speed of magnitude $v_{i,max}$. The saturation function $sat$ is defined as:
\begin{equation}
sat
    \left(
    \frac{s_i}{\xi_i\cdot tol_i}\right)=\begin{cases}
1, & \text{ if } \left(
    \frac{s_i}{\xi_i\cdot tol_i}\right)>1\\
\left(
    \frac{s_i}{\xi_i\cdot tol_i}\right), & \text{ if } -1\leq \left(
    \frac{s_i}{\xi_i\cdot tol_i}\right) \leq 1\\
-1, & \text{ if } \left(
    \frac{s_i}{\xi_i\cdot tol_i}\right)<-1
\end{cases}
\end{equation}

Since both cruising speed and altitude control are critical for autonomy and safety, steady-state errors need to be addressed. Indeed, damping effects, as well as changes in weight and buoyancy, will induce slower velocities than $v_{max}$. This can be solved using a recursive Sliding Moving Average (SMA) of the applied forces ($\tau^*_{SMA}$)  to correct them in real time. This approach requires minimal tuning (compared to PID control), and the sliding time window can be selected to achieve a more reactive controller with shorter time selections. The applied force $\tau^*$ is then defined as:
\begin{equation}
    \tau^*_i=\tau_i+\tau^*_{i,SMA}
\end{equation}
and its application can be represented in the following scheme:

\begin{figure}[H]
    \centering
    \includegraphics[width=0.7\textwidth]{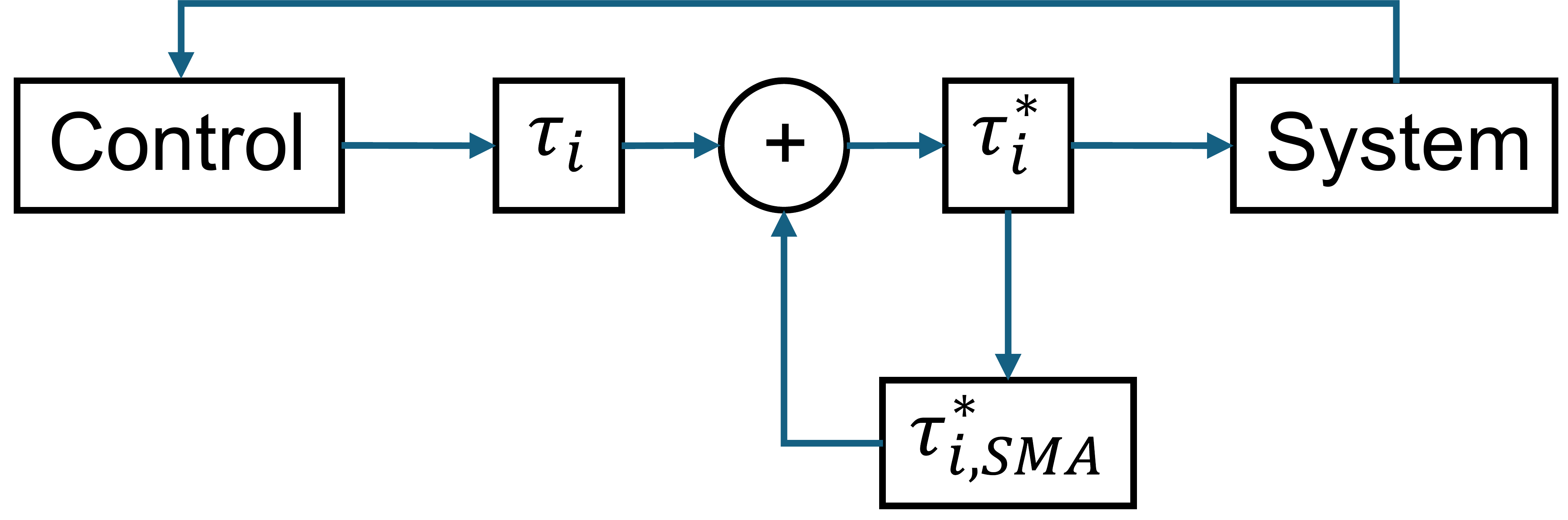}
    \caption{Control scheme, the SMA of the applied force is reintroduced in the controller before being applied to the system to reduce steady-state errors.}
    \label{fig:control_scheme}
\end{figure}

In the end, the value of $F_i$ is determined by the capacity of the actuators and $tol_i$ is a manually defined parameter that can be understood as the transition region between SMC and PD control. As for the value of $v_{max}$, it is derived from an optimization process that aims at an optimal flight of the robot, considering net weight (weight minus buoyancy), damping effects, and power consumption for its exploration mission.





\section{Use case: cave exploration as an artistic manifestation}\label{sec12}


\subsection{The cave}
\label{section:Cave}

As a use case to demonstrate the adaptability of the parametric design pipeline, we aim to fly inside a newly discovered urban cave located in Montreal. This cave is unique for its glaciotectonic geological features~\citep{pierre-etienne_caverne_2018}. The rock and fossils date back to an early ocean over 450 million years old. Fifteen thousand years ago, gigantic glaciers caused the widening of cracks in the limestone rock, creating the cave. Initially 35 meters long (the “Public Part”), a much deeper section of 275 meters was discovered in 2017~\citep{cloutier_dimmenses_2017, forget_caverne_2019}. This new section consists of a cavity (“Radiesthesia Part”) and access to a water level (“Aquatic Part”). The temperature remains constant at 12 °C, while the water temperature ranges between 6 °C and 9 °C, necessitating the use of wetsuits and safety harnesses through several technical passages, where wedges and ladders are installed at different levels of the cave.

\begin{figure}[H]
\centering
\includegraphics[width=0.9\textwidth]{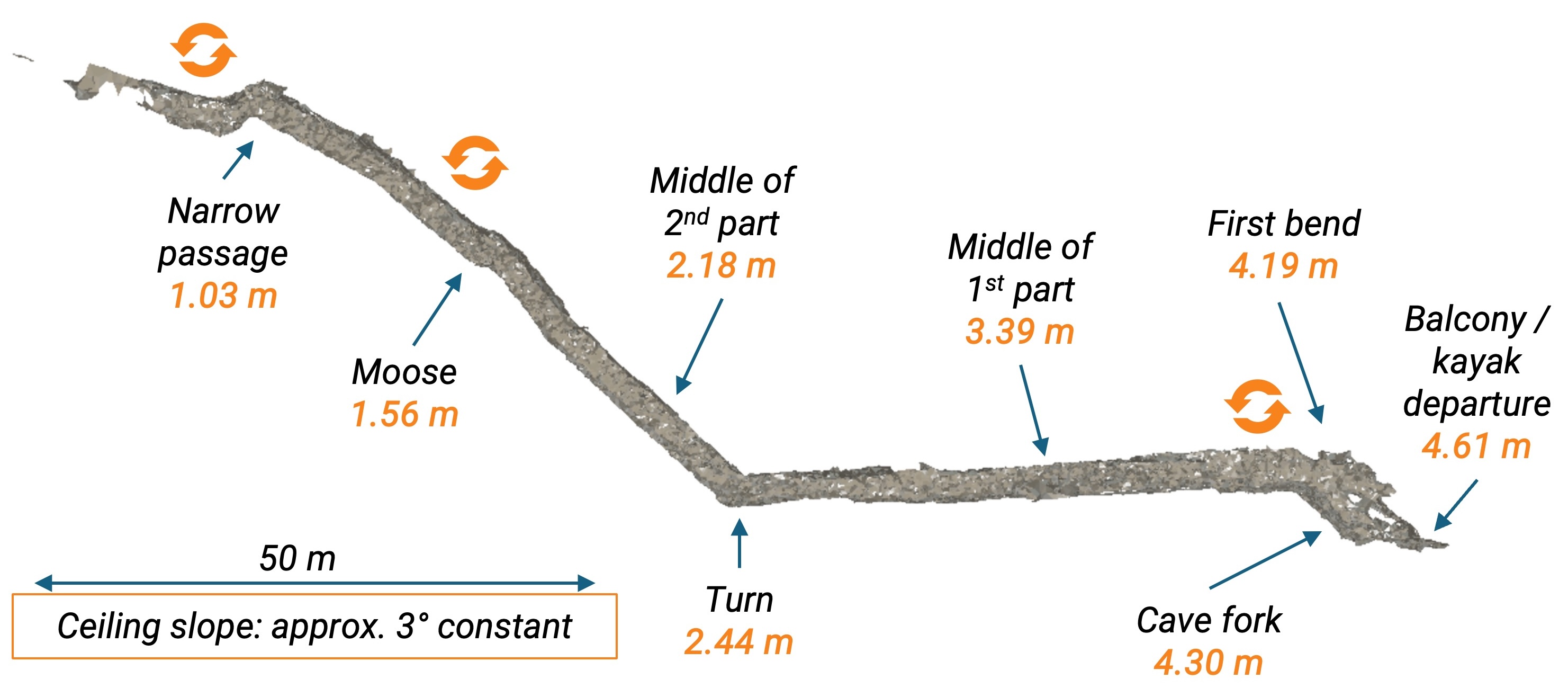}
\caption{Saint-Léonard “Aquatic part”. The three places where U-turns are possible are indicated by circular arrows. Ceiling heights are indicated at various checkpoints. The water level varies according to season and weather. }
\label{fig:cave_map}
\end{figure}

The exploration zone for the CAVERNAUTE is the “Aquatic Part ”, which constitutes the majority of the cave in terms of area. It is a long, flooded corridor about 0.8 to 2.5 meters wide, as shown in Fig.~\ref{fig:cave_map}. The ceiling starts at less than 5 meters high and gradually slopes down to the water level. Movement within that part of the cave requires small kayaks, as there are no areas to progress on foot. Getting onboard the kayak is quite technical from a narrow balcony, where one descends along a ladder directly to the kayak. Along the way in the cave, turning around with the kayak is only possible in three places, necessitating backward movement and leaning on the walls when the passage is too narrow for the paddles. 

\begin{figure}[H]
\centering
\includegraphics[width=0.5\textwidth]{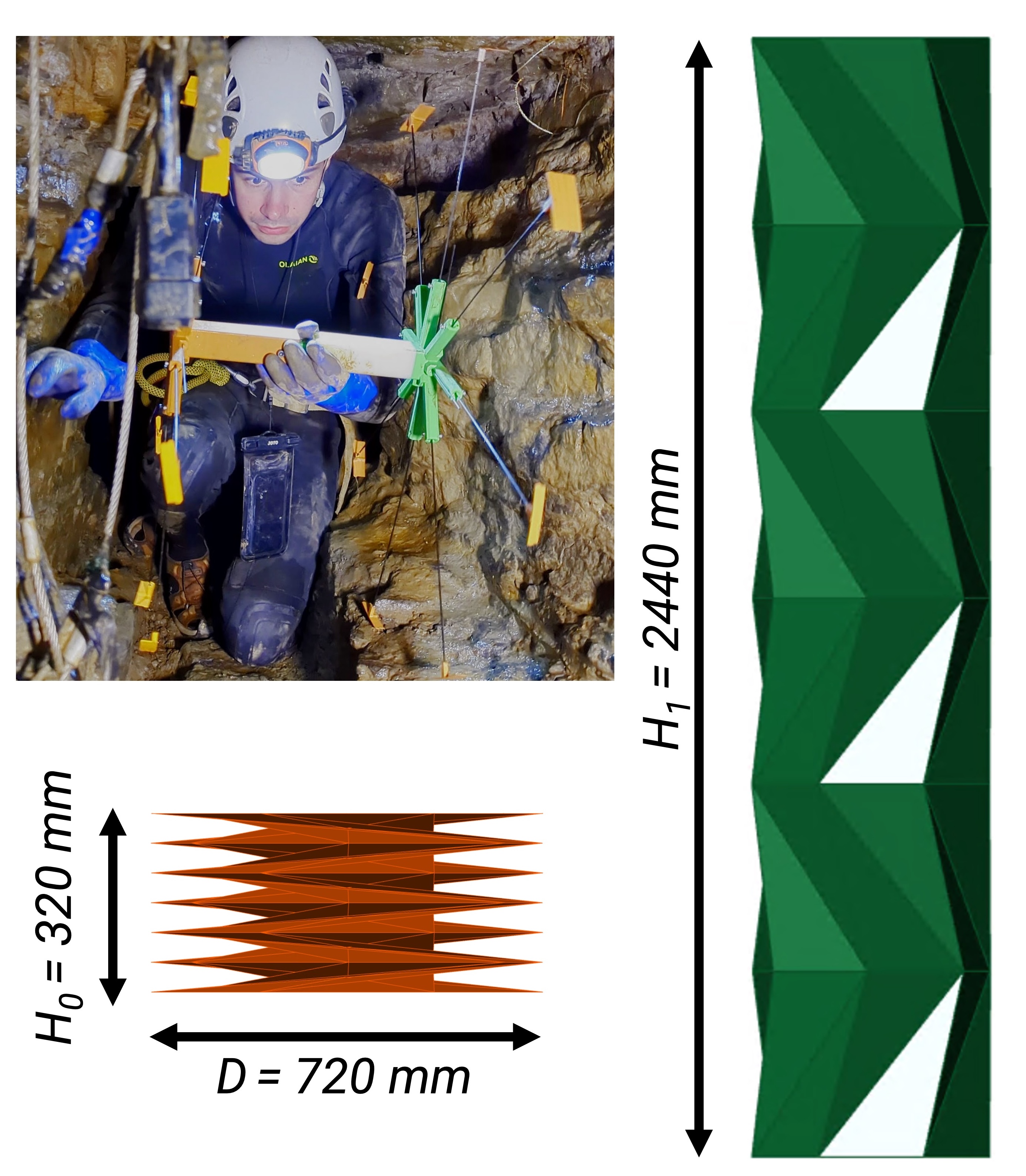}
\caption{Geometric dimensions constrained by the cave. The folded height $H_0$ and diameter $D$ of the CAVERNAUTE are dictated by the narrow passage to the exploration zone. The photo shows the tool used to measure and validate the ability to pass with the CAVERNAUTE. }
\label{fig:cave}
\end{figure}

Among all the input parameters presented in Table~\ref{tab:number}, three are related to geometry and determined by the constraints of the cave dimensions. Figure~\ref{fig:cave} details these three critical geometric data points. The first is the height of the unfolded CAVERNAUTE, which is bounded by the minimum vertical clearance for the airship to fly. This height was measured with a laser rangefinder. The other two parameters are the height and diameter of the folded CAVERNAUTE, both bounded by the narrowest point of passage to reach the exploration area. This critical point here is a thin corridor between the “Public Part” and the large cavity of the “Radiesthesia Part”. The narrowing was measured using a specially designed tool shown in Fig.~\ref{fig:cave}. It is a cylinder whose diameter and height are adjustable to be able to test the footprint of the folded CAVERNAUTE.

\subsection{Mapping the cave}


The mapping of the cave was made by our team in three sessions of about 5 hours each on different dates, using a custom handheld device with two or three LiDAR devices, an inertial measurement unit and a microcomputer running Linux with ROS. 

In the mapping process, a team of two traverses through the different sections of the cave. One holding the device and suitcase, and the other as a support to help when climbing or descending through certain sections of the cave. In the “Aquatic Part”, the person holding the device goes inside the kayak and the support member goes swimming and pulling/pushing the kayak. This is done to secure the smooth motion of the kayak and to avoid the occlusion of the LiDARs.

Three different LiDARs were used: an Ouster OS0\footnote{\href{https://ouster.com/products/hardware/os0-lidar-sensor}{https://ouster.com/products/hardware/os0-lidar-sensor}}, a Velodyne VLP-32MR LiDAR, and a RoboSense Helios LiDAR-16\footnote{\href{https://www.robosense.ai/en/rslidar/RS-Helios}{https://www.robosense.ai/en/rslidar/RS-Helios}}, working together with a VectorNav's VN-100\footnote{\href{https://www.vectornav.com/products/detail/vn-100}{https://www.vectornav.com/products/detail/vn-100}}. The devices were connected to an NVIDIA Orin Nano computer with Ubuntu 20, leveraging ROS and NorLab's ICP mapper\footnote{\href{https://github.com/norlab-ulaval/norlab\_icp\_mapper}{https://github.com/norlab-ulaval/norlab\_icp\_mapper}}.

After obtaining the point clouds, the different sections are stitched together to achieve a single combined point cloud of the cave, which we use as a reference to plan the flight of CAVERNAUTE.

\begin{figure}[H]
\centering
\includegraphics[width=0.5\textwidth]{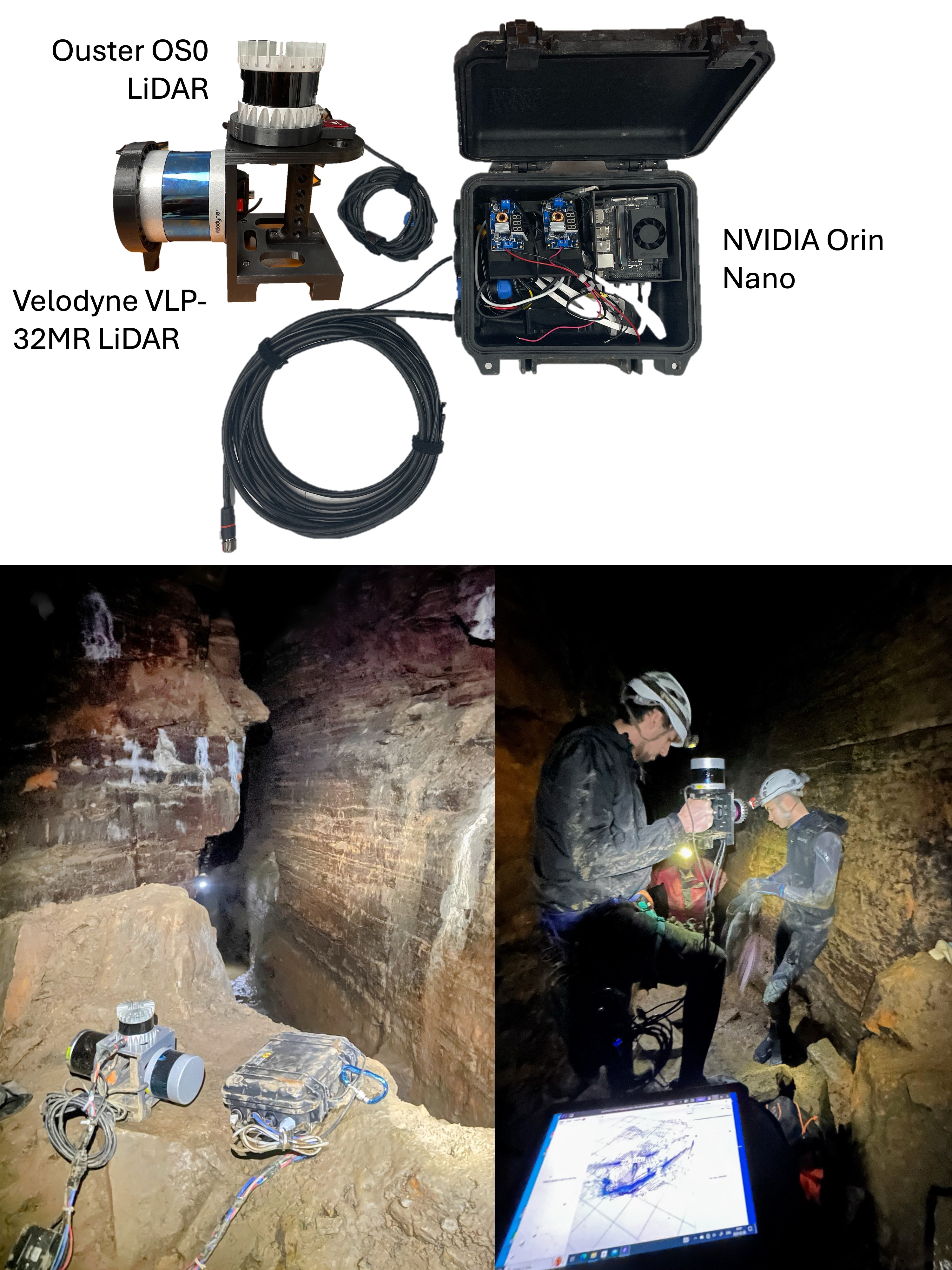}
\caption{On top, a photo of the device with two LiDARs. Bottom left, is an in-cavern photo of the device with three LiDARs, and the bottom right is in use with remote SSH access with a rugged tablet. }
\label{fig:map}
\end{figure}

\begin{figure}[H]
\centering
\includegraphics[width=1\textwidth]{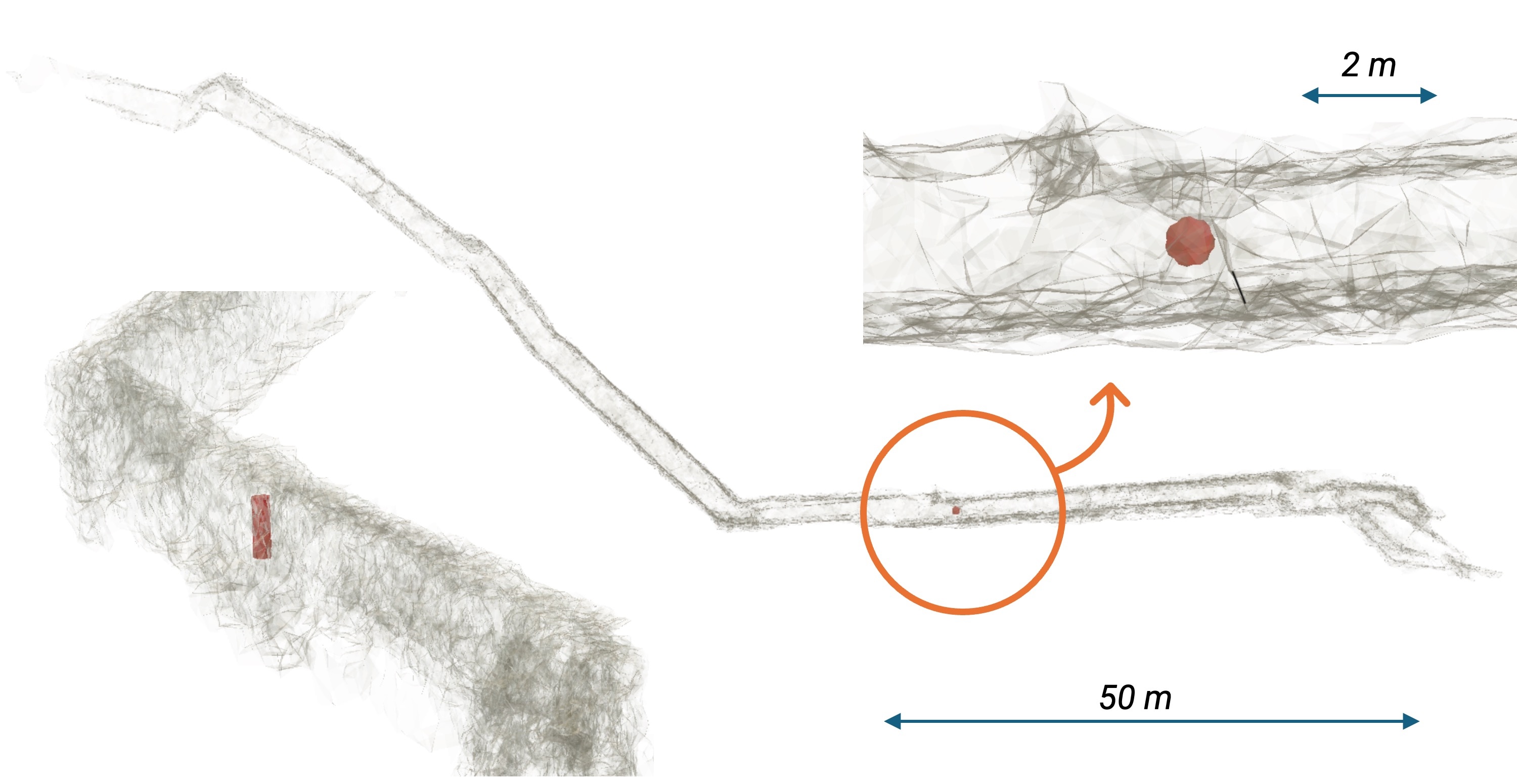}
\caption{CAVERNAUTE footprint in the exploration area mapped by the team}
\label{fig:cvnt_cave}
\end{figure}

Thanks to our preliminary mapping work, we can determine the flight zone for the CAVERNAUTE and thus the maximum height of the module to design as well as the narrowest section to transport the robotic system and its equipment. Figure~\ref{fig:cvnt_cave} shows the 2D model of the envelope in the “aquatic” zone.  

\subsection{Applied parametric design}

The deployment of the CAVERNAUTE has an artistic scope that has imposed several choices that can sometimes go against technical performance, but which has its importance for public visibility. A visual specification therefore brings several criteria subject to aesthetics. The first major criterion is the shape of the airship. A flexible airship has a shape defined by the pressure of the internal gas. This gives it an ovoid shape. Here, the adoption of a rigid structure gives the choice to create a non-conventional shape for the airship. Always in an artistic approach, it was decided to adopt a polyhedral shape: sharp edges that break the usual curves for a blimp to catch the interest of the public, who do not expect to see such a pure geometric shape able to float. The second criterion is that the airship must fly vertically, again with the aim of surprising. The impact on performance is the increase in drag on the axes of movement in the horizontal plane.

Next, the challenging flight environment imposes several constraints. The airship must be able to withstand impacts with the narrow walls of the cave. These same walls must not be damaged in the interest of conservation: the propellers will be protected by guards, and the sharp edges equipped with shock-absorbing structures. The speed of characterization of the structures was set at 2 m/s according to the maximum speeds observed in the literature for an indoor aerial system~\citep{nitta_visual_2017, tao_design_2020}. As the exploration area is located above water, the airship must be able to land on water in case of a problem without the electronics being submerged in water.

Finally, due to the difficulty of the mission, all structural elements must be replaceable in case of breakage during the planned 2-week deployment or to evolve the shape configuration. For the airship's edges, 2 mm carbon fiber tubes with a thickness of 0.5 mm were selected. These tubes have a tensile strength of 1200 MPa and a flexural strength of 615 MPa, with a linear density of m\textsub{CT} = 3.76~g/m.


The design window for the three parameters to optimize are n = $\llbracket 3;10 \rrbracket$, m = $\llbracket 2;10 \rrbracket$ and $\lambda$ = $[0.50;0.90]$. The requirement to have $\lambda$ less than 0.9 imposes the twisting of the structure so that it appears aesthetically more complex than a simple extruded polygon ($\lambda = 1$).This generates a total of 2880 configurations for investigation. 

\begin{figure}[H]
\centering
\includegraphics[width=0.95\textwidth]{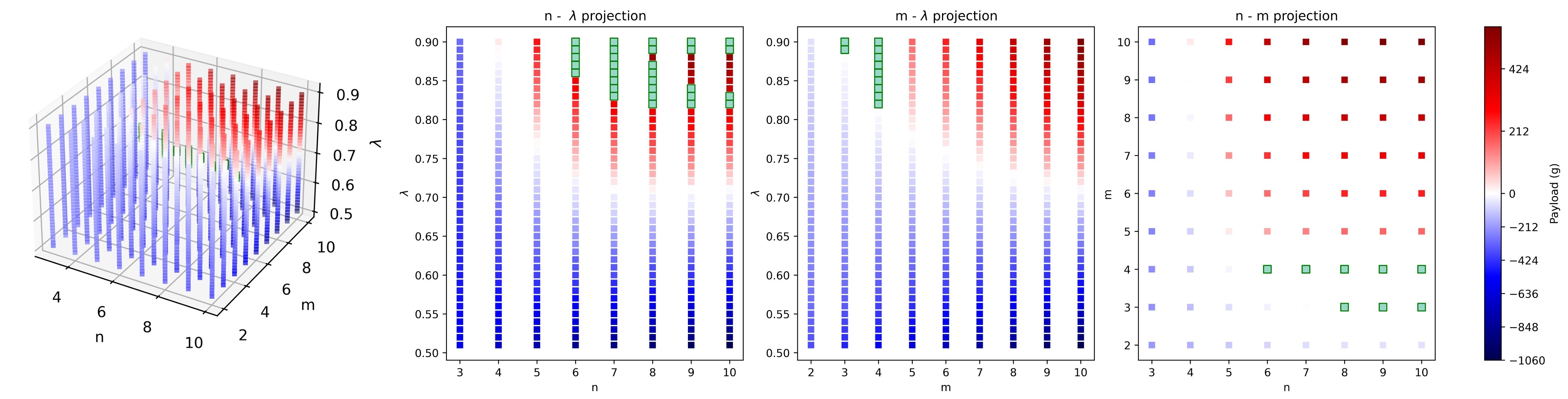}
\caption{Design window map and projection. Configurations that float (Payload $>$ 0) and respect the geometric constraints of the cavern are highlighted in green. }
\label{fig:3D}
\end{figure}

Figure~\ref{fig:3D} shows the map of these 2882 configurations based on the three parameters. The white area shows the floatability boundary. Naturally, the maximum payload is obtained for the largest n, m and $\lambda$ because we are geometrically getting closer to a cylinder, which maximizes the volume and therefore the payload. On the other hand, the geometric criteria drawn from the cavern eliminate numerous configurations. The valid configurations are therefore very close to the payload limits (white area). We also notice that m $<$ 5 because the unfolded height affects the number of possible Kresling segments. We see that the number of sides maximizes the volume but has an impact on the weight of the exoskeleton, which eliminates configurations.

\begin{figure}[H]
\centering
\includegraphics[width=0.6\textwidth]{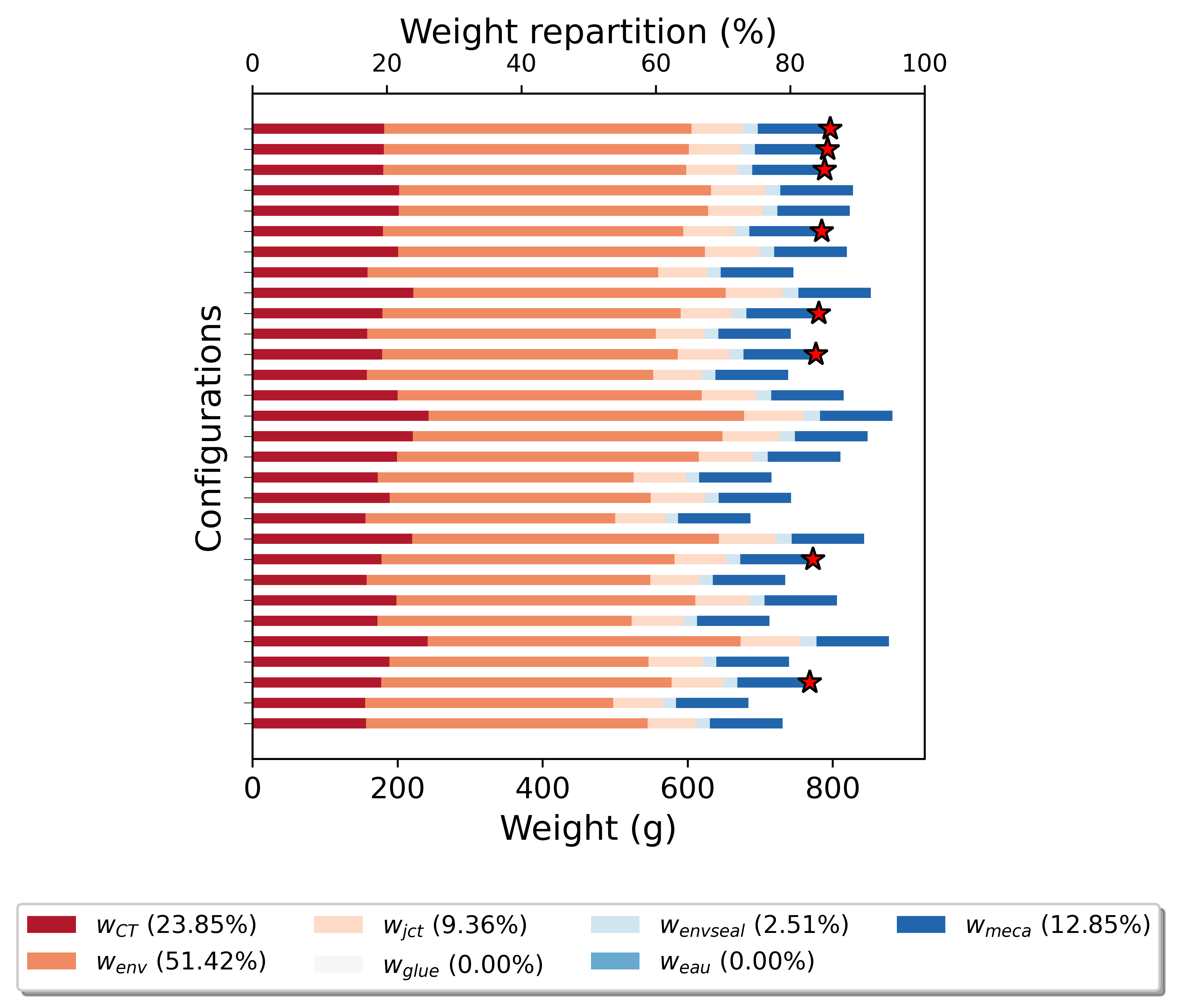}
\caption{Display of the weight distribution for the 30 realistic configurations. The red stars represent the best choice of the pair n, m. Here, respectively, n = 7 and m = 4.}
\label{fig:configuration}
\end{figure}

Figure~\ref{fig:configuration} shows the selection of the 30 configurations that validate the criterion payload $>$ 0 (the green squares highlighted in Fig.~\ref{fig:3D}). The weight distribution is displayed. On average, the weight of the envelope represents more than 50\% of the total weight. Therefore, the choice of the envelope material is crucial. The exoskeleton with carbon tubes and joints represents a total of 33\%. Finally, mechatronics occupies less than 13\%. This is the major drawback of a rigid airship: choosing to increase the resilience of the aerial system at the costs of the carrying capacity for sensors, the battery, etc.

\begin{figure}[H]
\centering
\includegraphics[width=0.8\textwidth]{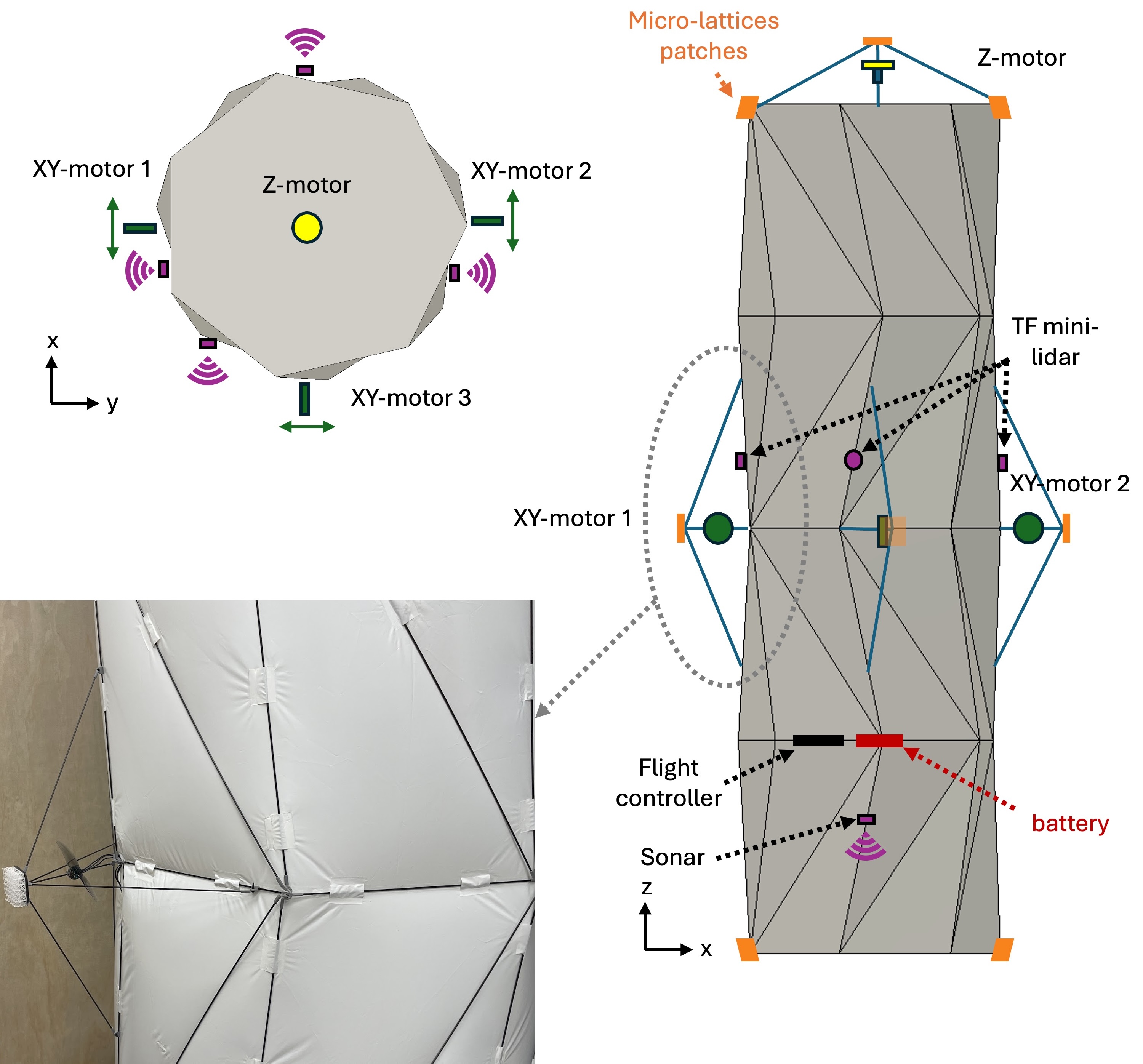}
\caption{Schematic of all CAVERNAUTE's design and motor position. The photo at bottom left shows the actual assembly of the side-mounted motor with carbon tube structure and micro-lattice patch. }
\label{fig:CVNT-design}
\end{figure}

To select the optimal configuration for CAVERNAUTE, it should be noted that the $\lambda$ parameter influences the twisting of the airship. This means that we can find a folding angle alpha for any lower $\lambda$. Thus, we are looking for the pair (n, m) with the highest number of occurrences among the valid configurations. Here, n = 7 and m = 4 is the optimal combination with $\lambda$ = [0.83; 0.9]. The extra-payload is between 1.1 g and 68 g. Concretely, we can adjust the extra-payload by twisting the $\lambda$ = 0.9 version of CAVERNAUTE. Table~\ref{tab:optimal} summarizes the data for the selected configuration. This ability to adjust the extra-payload can be very useful for precisely balancing the airship at the time of deployment. There is a risk of having water projections, sensor changes or exoskeleton configuration changes. Thus, this flexibility is a feature of CAVERNAUTE that facilitates the use of the aircraft. Figure~\ref{fig:CVNT-design} shows the geometry and the mechatronic component position on the optimized shape n = 7, m = 4 and $\lambda$ = 0.9

\begin{table}[ht!]
\caption{Extra-payload, height and bending angle relative to $\lambda$ = 0.9 for the selected configurations.}
\label{tab:optimal}
\begin{tabular}{cccc}
$\boldsymbol{\lambda}$ & \textbf{Extra-payload (g)} & \textbf{Height (mm)} & $\boldsymbol{\alpha_0}$ \textbf{(deg)} \\
0.83            & 1.1                        & 2265                 & 22                    \\
0.84            & 8.6                        & 2288                 & 21                    \\
0.85            & 25.1                       & 2325                 & 19                    \\
0.86            & 32.3                       & 2346                 & 18                    \\
0.87            & 39.3                       & 2366                 & 17                    \\
0.88            & 55                         & 2400                 & 15                    \\
0.89            & 61.5                       & 2418                 & 14                    \\
0.9             & 68                         & 2436                 & 13                   
\end{tabular}
\end{table}

\subsection{Benefits of origami properties for an indoor airship}

The use of an exoskeleton as a structure stiffening the envelope of an airship is new at this scale of airship and brings a succession of unprecedented advantages compared to large rigid airships and small blimps (soft envelope).

The exoskeleton provides resistance to abrasion because it is external. The thin envelope is protected from the majority of contacts due to its concave shape.

This bistability presented in Section~\ref{sec:kresling} has multiple interests. First, during transport, there are no constraints applied to the exoskeleton to keep it folded. Then, during inflation, each of the 4 segments that constitute the CAVERNAUTE can be filled and unfolded one after the other. This process limits air contamination in the helium-filled envelope. Once operational, in case of puncture of the envelope, the aerial system will be little affected by the damage. Indeed, unlike a soft blimp, the shape of the airship does not need to be maintained by the internal pressure of the carrier gas. So \textit{(1)} the airship can be inflated to a pressure close to atmospheric pressure and \textit{(2)} if the airship is inflated to a higher pressure, in case of a puncture, the pressure will balance to atmospheric pressure and the leak will be almost stopped.

Derived from bistability and with $\lambda$ close to 1 to maximize energy potential (Fig.~\ref{fig:energy}), another property of the Kresling pattern is its excellent shock resilience~\citep{yasuuda_origami-based_2019, oneil_energy_2023}. Here, in addition to the high-performance absorption capabilities of micro-lattices, the exoskeleton of the CAVERNAUTE provides a second level of absorption. Thanks to the rigid exoskeleton, the inertia of the heaviest objects that make up the airship such as the battery, sensors, does not affect the envelope. A soft airship risks tears because the masses are suspended with a gondola. With the CAVERNAUTE, the loads are transmitted inside the exoskeleton and are isolated from the fragile envelope.

Finally, the volume expansion ratio is:
\begin{equation}
    \frac{V_{deployed}}{V_{folded}} = \frac{0.825}{0.0417}[m^3] = 19.78
\end{equation}

This very high ratio allows the CAVERNAUTE to fit in a bag to get through complex passages in caving to access exploration areas. Benefiting from rigidity is therefore not a problem for remote scientific missions with this compactness.

\subsection{Controller Optimization}
\label{section:control_optimization}

In order to roughly estimate the energy required for the CAVERNAUTE's mission, we multiply the total power required by the time to travel the distance of the cave's “Aquatic Part”. We can then optimize the controller's parameters with fewer degrees of freedom, namely: the global path error, the maximum velocity and the time to complete its journey. Beforehand, the consumption of the Cavernaute is estimated component by component from measurements or using the manufacturers datasheets.

\begin{figure}[H]
\centering
\includegraphics[width=0.75\textwidth]{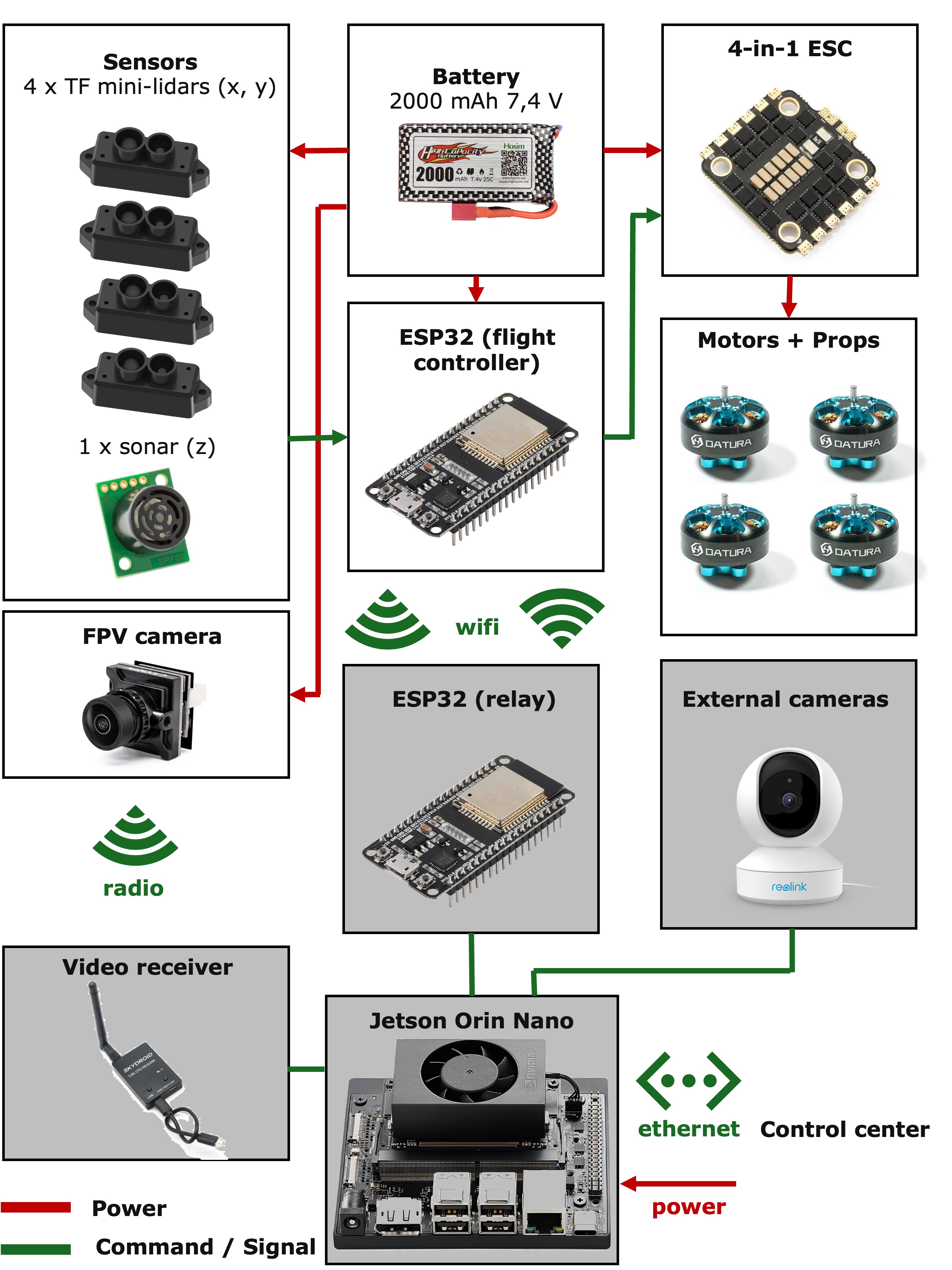}
\caption{Schematic of all CAVERNAUTE's mechatronic: components with gray backgrounds are on the ground. The external camera is only for retransmitting the artistic installation.}
\label{fig:mechatronic}
\end{figure}

We are using a Hosim RC Cars Battery of 7.4 V with 2Ah, thus delivering a capacity of 14.8 Wh used to power 4 Benewake Micro TFmini I2C LiDARs, one FPV Caddx Baby Ratel 2 camera, the ESP32 controller development kit and a small SRF02 I2c sonar. The power consumption of each device is presented in Table \ref{tab:power_devices}. Figure~\ref{fig:mechatronic} shows the schematic of all mechatronic components. 

\begin{table}[h]
\caption{Power consumption for each of the devices used for CAVERNAUTE}
\label{tab:power_devices}
\begin{tabular}{lrrr}
\textbf{Part} & \multicolumn{1}{l}{\textbf{Power Consumption {[}W{]}}} & \multicolumn{1}{l}{\textbf{Quantity}} & \multicolumn{1}{l}{\textbf{Total Power {[}W{]}}} \\
TFmini LiDARs & 0.7\footnotemark & 4 & 2.8 \\
Sonar & 0.015 & 1 & 0.015 \\
Video Transmitter & 0.5 & 1 & 0.5 \\
Camera & $<$1.2 & 1 & $<$1.2 \\
 & \multicolumn{1}{l}{} & \multicolumn{1}{l}{\textbf{Total}} & $<$4.515
\end{tabular}
\end{table}
\footnotetext{We measured 0.25W when pulling data from the LiDARs at 10Hz, but we kept the datasheet's value to be conservative.}
The motors selected for our design are EMAX RS1606 3300 KV that we characterized using a TytoRobotics Series 1520 Thrust Stand \footnote{\href{https://www.tytorobotics.com/pages/series-1520}{https://www.tytorobotics.com/pages/series-1520}}. We ended up with a quadratic regression between the force (in Newtons) and the power (in Watts) for the motors as:
\begin{equation}
    P=3.347F^2-25.857F+1.69
\end{equation}
By design, we decide on having a slight heavier-than-air prototype. This is a common design strategy to avoid fly-aways, however, in our case, the rational is rather to protect the envelop from the ceiling's stalactites, plus our design can float in the cave's water without harm (see Fig.\ref{fig:CVNT-design}). We consider the net-weight (weight minus buoyancy) is approximately 0.02 kgf, which means a consumption of 6.9 Watts with a single (Z-oriented) motor.

\begin{figure}[H]
\centering
\includegraphics[width=0.55\textwidth]{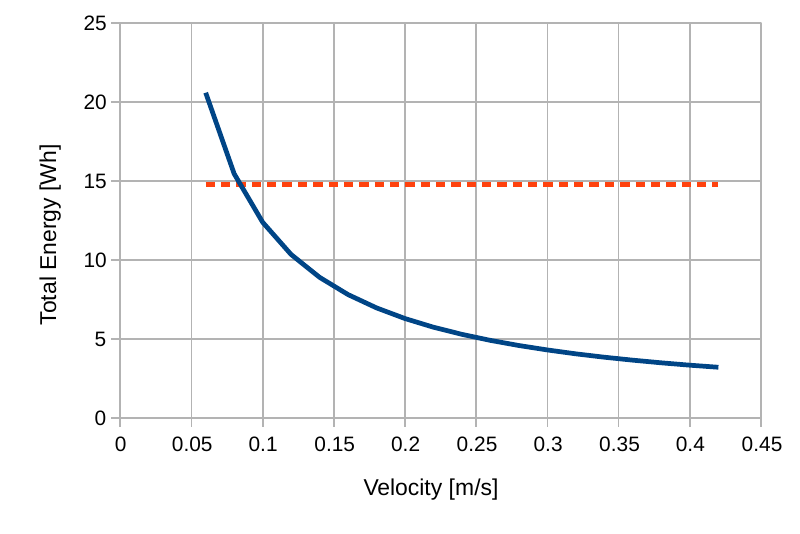}
\caption{Relation between flight velocity and estimated total energy consumption (in blue). The total available energy provided by the battery is presented in the dashed red line.}
\label{fig:velocity_energy}
\end{figure}

For translation motion, we need to consider damping effects, which for an airship, even at low velocity, consists in the aerodynamic drag. We approximate the cross-section of the CAVERNAUTE as a circle of area equal to 0.41 m\textsuperscript{2}, leading to a drag coefficient of about 1.2~\citep{bruschi_drag_2003} at slow velocity (and low Reynolds numbers). The air inside the cavern is of the same composition and density as atmospheric air at 12 °C: 1.231 kg/m\textsuperscript{3}. The dampening force is calculated as:
\begin{equation}
    F_D=\frac{1}{2}C_D\rho v^2 A
\end{equation}
We integrate the dampening force to maintain the cruising speed, and estimate the power as:
\begin{equation}
    P=3.347\left(\frac{1}{2}C_D\rho v^2 A\right)^2-25.857\left(\frac{1}{2}C_D\rho v^2 A\right)+1.69
\end{equation}
With a total electronic consumption equal to 11.4 W in hovering mode (sensors plus Z-motor), and by setting the two motors for the forward direction at cruising speed, the relation between velocity  and power budget to travel 300 meters inside the cave (conservative distance) is illustrated in Fig.~\ref{fig:velocity_energy} for various cruising velocities.
From Fig. \ref{fig:velocity_energy} we extract a minimum flying speed of 0.08 m/s, which corresponds to 62 minutes of flight. More efficient flights require higher speeds. Nevertheless, this first approximation does not consider the energy required to maintain the airship far from the cave's walls (Y-axis collision avoidance).

\subsection{Simulation}

To demonstrate our controller, we present a simple motion: starting from the ground up to maintain an altitude of 1 meter, then moving forward a distance of 2 meters at 0.15 m/s. This test is repeated under three conditions: one without the SMA term, and then with two with different duration (1 and 2 seconds) for the SMA term.

Our custom simulation stack solves the dynamics equations in Python at 500 Hz, and integrate the different control modules using ROS, with each node running at a realistic frequency of 40 Hz. We also used Gazebo to visualize the motion, but Gazebo's solver was not used.

We consider the added mass terms and drag for the forward direction $X$ and the altitude $Z$. For $Z$, the approximated drag coefficient of the flat circular face is approximately 0.9, and the added mass was calculated as the contribution of the added mass of rectangular sections in rotation, following~\cite{korotkin_added_2009}, equal to 0.29 kg. For $X$, the added mass obtained was equal to 1.165 kg, and the drag parameters are already presented in Section \ref{section:control_optimization}.

\begin{figure}[H]
\centering
\includegraphics[width=0.45\textwidth]{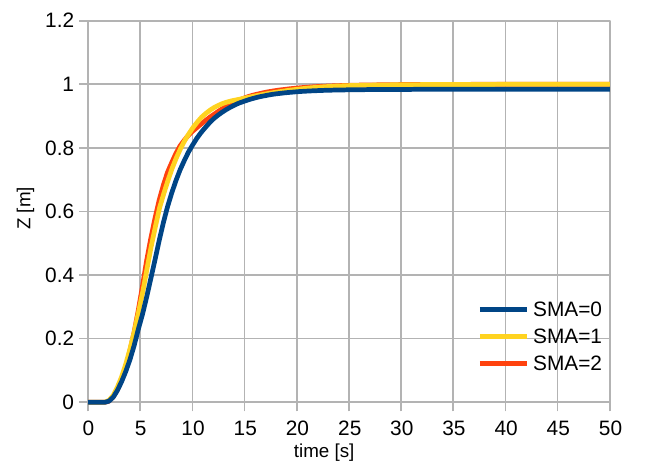}
\includegraphics[width=0.45\textwidth]{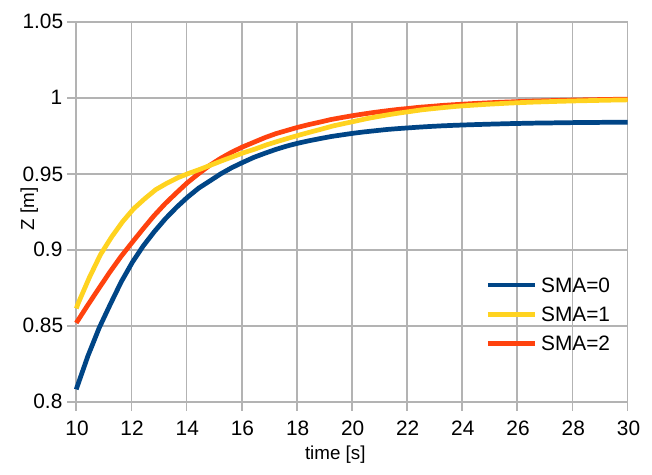}
\includegraphics[width=0.45\textwidth]{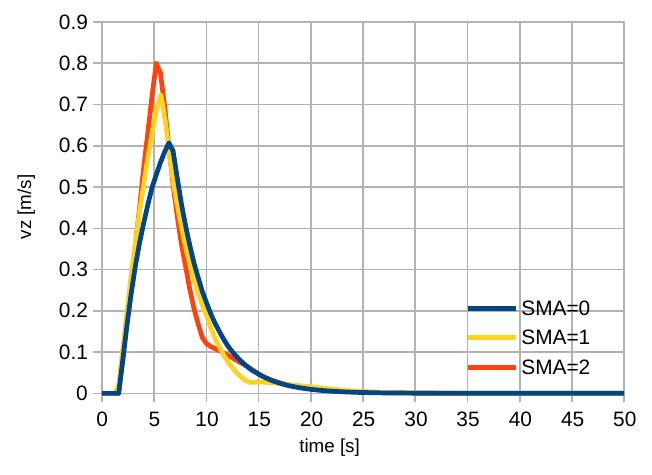}
\caption{Motion in the $Z$-axis. The top graphs present the position, for the different SMA durations. The bottom graph presents the velocity achieved ($v_z$).}
\label{fig:sim_Z}
\end{figure}

We set $F_x=F_z=1.25$ N to consider the maximum actuation of the motors, $v_{x,max}=0.15$ m/s as the selected cruising velocity, $v_{z,max}=1$ m/s to allow for faster motion in the $Z$ axis since we need to achieve and maintain the desired height quickly, and using $tol_x=tol_z=0.1$ as the threshold between SMC and PD control.

Figure \ref{fig:sim_Z} presents the first 50 seconds of the complete motion in the $Z$ axis, as the Cavernaute achieves its 1 meter target altitude. The SMA terms are correcting the small steady state error.

\begin{figure}[H]
\centering
\includegraphics[width=0.45\textwidth]{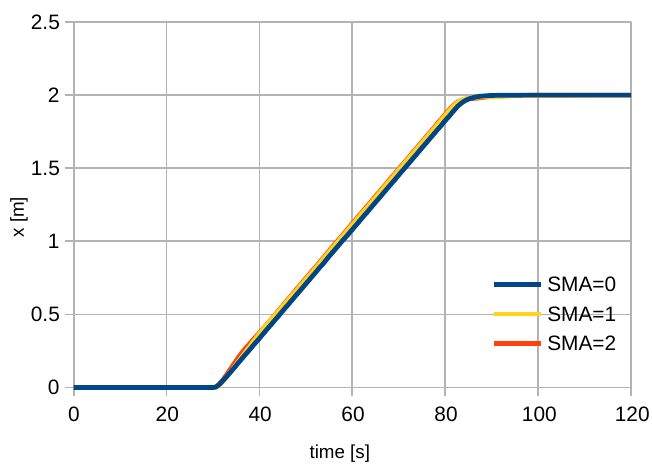}
\includegraphics[width=0.45\textwidth]{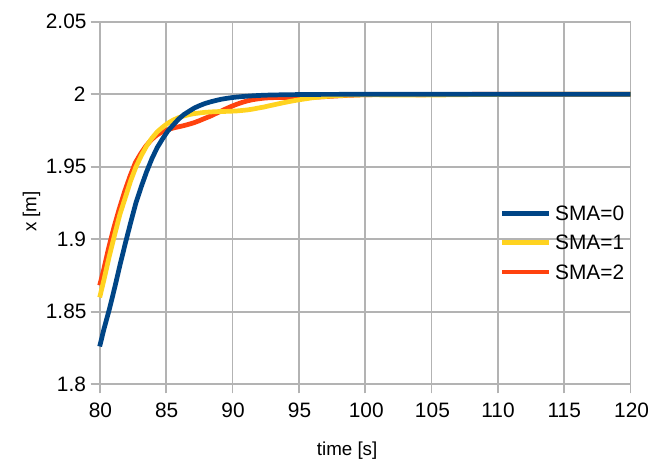}
\caption{Motion in the $X$-axis. The two graphs present the position to achieve the desired position of 2 m, and the effect of the SMA terms.}
\label{fig:sim_X1}
\end{figure}

\begin{figure}[H]
\centering
\includegraphics[width=0.45\textwidth]{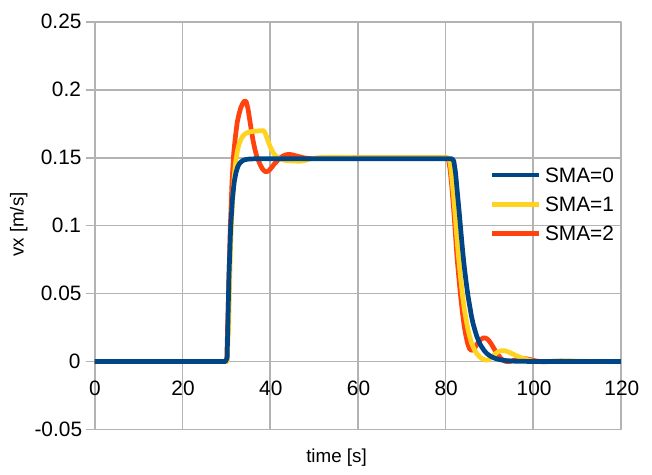}
\includegraphics[width=0.45\textwidth]{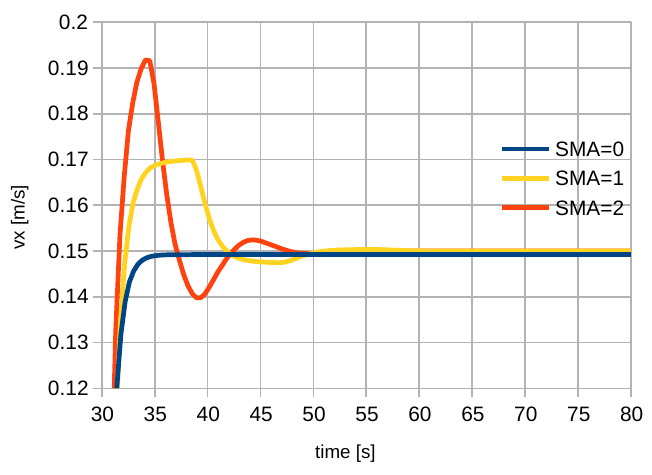}
\caption{Motion in the $X$-axis. The two graphs present the velocity ($vx$) while moving towards the position of 2 m, presenting the effect of the SMA terms.}
\label{fig:sim_X2}
\end{figure}

In Fig. \ref{fig:sim_X1} and \ref{fig:sim_X2}, we observe that all the controllers were capable to achieve the desired position of 2 m. Nevertheless, there is a slight difference in the velocity, where both SMA durations were able to achieve a cruising motion of 1.5 m/s. We can attribute this difference to the small drag force produced when moving at the slow speed of 0.15 m/s.

In all the cases, the simplicity of the methodology to tune the motions based on the speed requirements, maximum desired force and desired tolerance was deemed satisfactory to be applied in our robot. And, while it might not be mandatory to include the SMA term for our conditions, we could observe that its inclusion in the controller made it possible to precisely correct the robot position and velocity to the desired requirements in the presence of a force (net-weight) and damping effects.

\section{Conclusion}

In conclusion, we have developed a small rigid airship with unique properties, leveraging origami and the Kresling pattern to stiffen the envelope while enabling access to narrow areas through folding capability. To optimize the intricate considerations of the airship, we have implemented a design, manufacturing, and assembly pipeline that integrates the constraints of the specifications, allowing for flexible and easy testing of new configurations. This approach has identified areas for improvement, particularly the need to lighten the envelope, which currently represents more than half of the weight—a key focus for future research.

Our work highlights several features enabled by combining origami with airship design, offering significant resilience from multiple perspectives. For robotic deployment in challenging environments, these features collectively reduce the risk of catastrophic failure that could compromise the mission.

The impact of this research extends to various applications such as structure inspection, security, and search and rescue operations. The ability to navigate confined spaces, coupled with extended flight duration and robust design, positions this airship as a valuable tool in these fields.

Future developments could further exploit these features, such as actuating the Kresling pattern to allow digital control of the extra-payload, thereby providing more precise control over buoyancy and enhancing the airship's adaptability and functionality in diverse scenarios.

\backmatter

\subsubsection*{Supplementary information}

The data that support the findings of this study are available from the corresponding author [LC], upon reasonable request.

\subsubsection*{Acknowledgements}
The authors would like to thank Raphael Desbiens, Guillaume Ricard, and Brendan Patience for their significant support in the prototyping, development and implementation of the CAVERNAUTE mechatronics.

The authors acknowledge the financial support of the NSERC CREATE UTILI program\footnote{https://carleton.ca/utili/}, and the FRQNT Team grant (\#283381). 

\subsubsection*{Conflict of interest} The authors declare that they have no conflict of interest.

\subsubsection*{Authors contributions}

L.C. was the project integrator for the airship aspects. He designed, manufactured and assembled the structure and mechanical components. He built the design pipeline described in this article. He wrote all the corresponding sections, the literature review and the conclusion. J.E.S.G. designed and wrote all the sections concerning airship navigation, control, and cave mapping. Both L.C. and J.E.S.G did the manual mapping and measurements of the cave as described in section \ref{section:Cave}. D.SO. provided the original concept for this work and participate to both the mechatronic and geometric designs. D.SO. and I.T. revised the manuscript and supervised the work. 









\begin{appendices}

\end{appendices}


\bibliography{references}

\end{document}